\documentclass[sigconf,nonacm,timestamp,authorversion,natbib]{acmart}

\AtBeginDocument{%
  }

\setcopyright{cc}
\copyrightyear{2025}

\usepackage{textcomp}

\usepackage{colortbl}

\usepackage[figuresright]{rotating}
\usepackage{xspace}

\usepackage{amsfonts}

\usepackage[english,ruled,vlined,linesnumbered]{algorithm2e}

\usepackage[capitalise]{cleveref}

\usepackage[printonlyused,smaller,nolist]{acronym}

\usepackage{mathtools}
\DeclarePairedDelimiter\set\{\}

\usepackage[shortlabels]{enumitem}

\usepackage{multirow}

\usepackage{makecell}

\usepackage{trimspaces}

\usepackage{wrapfig}

\usepackage{dirtytalk}

\usepackage{checkend}

\usepackage{fixfoot}

\usepackage{hyperref}

\newcommand\Double{\KwSty{Double}}
\newcommand\Int{\KwSty{Int}}

\newcommand\String{\KwSty{String}}

\SetKwFunction{CreateStringBuilder}{StringBuilder}

\SetKwFunction{CreateDictionary}{Dictionary}

\NewDocumentCommand\NormalBrackets{ o }{\ignorespaces\FuncSty{(}\IfValueTF{#1}{#1}{\,}\FuncSty{)}}
\NewDocumentCommand\CurlyBrackets{ o }{\ignorespaces\FuncSty{\{}\IfValueTF{#1}{#1}{\,}\FuncSty{\}}}
\NewDocumentCommand\SquareBrackets{ o }{\ignorespaces\FuncSty{\string[}\IfValueTF{#1}{#1}{\,}\FuncSty{\string]}}

\NewDocumentCommand\List{ o }{\KwSty{List}\IfValueT{#1}{\SquareBrackets[#1]}}
\NewDocumentCommand\Set{ o }{\KwSty{Set}\IfValueT{#1}{\SquareBrackets[#1]}}
\NewDocumentCommand\HashSet{ o }{\KwSty{HashSet}\IfValueT{#1}{\SquareBrackets[#1]}}
\NewDocumentCommand\Map{ m m }{\KwSty{Map}\SquareBrackets[#1, #2]}

\NewDocumentCommand\Tuple{ s o }{\IfBooleanF{#1}{\KwSty{Tuple}}\IfValueT{#2}{\NormalBrackets[#2]}}

\SetKwData{First}{first}
\SetKwData{Second}{second}

\newcommand\EmptyList{\SquareBrackets}
\newcommand\EmptySet{\CurlyBrackets}
\newcommand\EmptyMap{\CurlyBrackets}
\newcommand\EmptyHashSet{\EmptySet}

\SetKwData{Keys}{keys}

\NewDocumentCommand\CreateTuple{>{\SplitList{,}}m}{%
    \let\privateitem\privateitema%
    \NormalBrackets[\ProcessList{#1}{\privateitem}]
}
\NewDocumentCommand\CreateList{>{\SplitList{,}}m}{%
    \let\privateitem\privateitema%
    \SquareBrackets[\ProcessList{#1}{\privateitem}]
}
\NewDocumentCommand\CreateSet{>{\SplitList{,}}m}{%
    \let\privateitem\privateitema%
    \CurlyBrackets[\ProcessList{#1}{\privateitem}]
}

\newcommand\privateitem[1]{--NO ONE EVER SEES THIS--}
\newcommand\privateitema[1]{#1\let\privateitem\privateitemb}
\newcommand\privateitemb[1]{\textnormal{,}#1}

\SetKwFunction{Add}{Add}
\SetKwFunction{GetOrCreate}{GetOrCreate}
\SetKwFunction{AddAll}{AddAll}
\SetKwFunction{SortAscending}{SortAscending}
\SetKwFunction{SortDescending}{SortDescending}
\SetKwFunction{Take}{Take}

\newcommand{\OfType}{$\!:\,$}

\newcommand{\Gets}{ $\gets$ }
\newcommand{\Deconstruct}{ $\gets$ }

\newcommand{\OpIn}{ \KwSty{in} }

\newcommand{\OpTo}{ \KwSty{to} }

\NewDocumentCommand{\OpGet}{ m }{\kern-3pt[#1\kern-3pt]}

\NewDocumentCommand{\OpInvoke}{ o }{\kern-3pt\IfValueTF{#1}{\NormalBrackets[#1\kern-3pt]}{\NormalBrackets}}

\SetKwProg{Fun}{fun}{ begin}{end}

\NewDocumentCommand\CallMember{ m }{\kern-1pt\textnormal{.}#1}

\NewDocumentCommand\AccessField{ m }{\kern-1pt\textnormal{.}#1}

\NewDocumentCommand\AccessProperty{ m }{\kern-1pt\textnormal{.}#1}

\NewDocumentCommand\AccessByReference{ m }{\textnormal{::}#1}

\NewDocumentCommand\Var{}{\KwSty{var} }
\NewDocumentCommand\Val{}{\KwSty{val} }

\NewDocumentCommand\Label{ s m }{\KwSty{\IfBooleanTF{#1}{#2@ }{#2:}}}
\NewDocumentCommand\Continue{ s o }{\KwSty{continue}\IfValueT{#2}{\KwSty{\IfBooleanTF{#1}{@#2}{ #2}}}}
\NewDocumentCommand\Break{ s o }{\KwSty{break}\IfValueT{#2}{\KwSty{\IfBooleanTF{#1}{@#2}{ #2}}}}
\NewDocumentCommand\GoTo{ s m }{\KwSty{goto}\IfValueT{#2}{\KwSty{\IfBooleanTF{#1}{@#2}{ #2}}}}

\NewDocumentCommand\TypeAlias{ m m }{\KwSty{typealias} #1 = #2}

\SetAlFnt{\small\sf} 
\SetKwSty{textnormal} 
\SetFuncArgSty{textnormal}
\SetArgSty{textnormal}
\SetAlgoNlRelativeSize{0}
\SetNlSkip{0em}
\SetNlSty{texttt}{[}{]}

\DeclareMathSymbol{\shortminus}{\mathbin}{AMSa}{"39}

\newenvironment{nospaceflalign*}
 {\setlength{\abovedisplayskip}{0pt}\setlength{\belowdisplayskip}{0pt}%
  \csname flalign*\endcsname}
 {\csname endflalign*\endcsname\ignorespacesafterend}

\newcommand\RankT{\Int}
\newcommand\ProbabilityT{\Double}

\SetKwData{Topic}{$\varphi_{k}$}
\SetKwData{TopicA}{$\varphi^{A}_{k}$}
\SetKwData{TopicB}{$\varphi^{B}_{k}$}
\SetKwData{TransTopicB}{$\varphi^{B}_{k}$}
\SetKwData{TransTopicBpp}{$\varphi^{B}_{k+1}$}
\SetKwData{TopicModel}{$T$}
\SetKwData{TopicModelA}{$T_A$}
\SetKwData{TopicModelB}{$T_B$}
\SetKwData{Vocabulary}{$V$}
\SetKwData{VocabularyA}{$V_A$}
\SetKwData{VocabularyB}{$V_B$}
\SetKwData{DictAB}{$\mathscr{D}^{A}_{B}$}
\SetKwData{DictBA}{$\mathscr{D}^{B}_{A}$}
\SetKwData{Epsilon}{$\epsilon$}
\SetKwData{Threshold}{$\xi$}
\SetKwData{NBest}{$n$}
\SetKwData{PartialTopics}{candidates}
\SetKwData{TopicMapping}{topicMapping}
\SetKwData{Lang}{language}
\SetKwData{Stemmer}{stemmer}
\SetKwData{Stopwords}{stopwwords}
\SetKwData{UseUencode}{unidecode}
\SetKwData{Lang}{lang}
\SetKwData{NULL}{NULL}
\SetKwData{ElectedCandidates}{elected}
\SetKwData{BestNCandidates}{best}
\SetKwData{MachineEpsilon}{$\varepsilon_{\text{Machine}}$}

\SetKwFunction{TranslateTopicModel}{TranslateTopicModel}
\SetKwFunction{Translate}{Translate}
\SetKwFunction{CreateVocabulary}{CreateVocabulary}
\SetKwFunction{CreateTopics}{CreateTopics}
\SetKwFunction{ToTopicWithHeuristic}{RunElection}
\SetKwFunction{ConvertToList}{ConvertToList}
\SetKwFunction{GetRank}{RankOf}
\SetKwFunction{CalculateScore}{CalculateScore}
\SetKwFunction{AddAllIfNotExists}{AddAllIgnoreDuplicates}
\SetKwFunction{AddAllHandleSameWord}{AddAllUnique}
\SetKwFunction{Rank}{Rank}
\SetKwFunction{Max}{Max}
\SetKwFunction{Min}{Min}
\SetKwFunction{MinOf}{MinOf}
\SetKwFunction{Average}{Avg}
\SetKwFunction{GeometricMean}{gMean}
\SetKwFunction{Tokenize}{TokenizeAndClean}
\SetKwFunction{PComSUM}{PComSUM}
\SetKwFunction{MinProbabilityIn}{MinProbabilityIn}
\SetKwFunction{Normalize}{Normalize}
\SetKwFunction{FitToVocabuarly}{FitToVocabuarly}
\SetKwFunction{Toppp}{Top3}
\SetKwFunction{score}{Score}

\newcommand\ds{\texttt{DAT}}
\newcommand\dsTrain{\texttt{DAT}\textsubscript{\texttt{train}}}
\newcommand\dsTest{\texttt{DAT}\textsubscript{\texttt{test}}}

\NewDocumentCommand\Voter{ s o }{\ignorespaces\IfBooleanF{#1}{\KwSty{Voter}}\IfValueT{#2}{\NormalBrackets[#2]}}

\NewDocumentCommand\Elected{ s o }{\ignorespaces\IfBooleanF{#1}{\KwSty{Elected}}\IfValueT{#2}{\NormalBrackets[#2]}}

\SetKwFunction{IsStopWord}{IsStopWord}
\SetKwFunction{IsPhrase}{IsPhrase}
\SetKwFunction{SplitPhrase}{SplitPhrase}
\SetKwFunction{Stem}{Stem}
\SetKwFunction{UniEncode}{FoldASCII}
\SetKwFunction{Build}{toString}
\SetKwFunction{LowerCase}{LowerCase}
\SetKwData{Word}{wordOrPhrase}
\SetKwData{Wordd}{word2}
\SetKwData{Result}{result}
\SetKwData{Phrase}{phrase}
\SetKwData{Stemmer}{stemmer}
\SetKwData{StemmerType}{Stemmer}
\SetKwData{Temp}{temp}
\SetKwData{Tokens}{tokens}
\SetKwData{Token}{token}

\SetKwData{Sentence}{sent}
\SetKwData{Dict}{$\mathscr{D}$}
\SetKwData{PhraseLength}{phraseLength}

\SetKwData{Stopwords}{stopwords}

\SetKwFunction{Tokenize}{Tokenize}
\SetKwFunction{TokenizeWithDict}{TokenizeWithPhrases}
\SetKwFunction{Calculate}{calculateScore}

\SetKwData{VoteScore}{score}
\SetKwData{ScoreModel}{VoteModel}

\NewDocumentCommand\dsEnwiki{s}{\IfBooleanTF{#1}{\textbf{enwiki}}{enwiki}}
\NewDocumentCommand\dsAligned{s}{\IfBooleanTF{#1}{\textbf{aligned}}{aligned}}
\NewDocumentCommand\dsNewsgroups{s}{\IfBooleanTF{#1}{\textbf{newsgroups}}{newsgroups}}

\newcommand{\Score}{\mathscr{S}}

\definecolor{Baseline}{HTML}{000000}
\definecolor{CombNOR}{HTML}{E50000}
\definecolor{CombGNOR}{HTML}{C20078}
\definecolor{CombGSUM}{HTML}{F97306}
\definecolor{CombSUM}{HTML}{15B01A}
\definecolor{CombSUMRR}{HTML}{029386}
\definecolor{CombRRPEN}{HTML}{0343DF}
\definecolor{RR}{HTML}{000080}

\definecolor{tmt_black}{HTML}{000000}
\definecolor{tmt_red}{HTML}{E50000}
\definecolor{tmt_magenta}{HTML}{C20078}
\definecolor{tmt_orange}{HTML}{F97306}
\definecolor{tmt_green}{HTML}{15B01A}
\definecolor{tmt_teal}{HTML}{029386}
\definecolor{tmt_blue}{HTML}{0343DF}
\definecolor{tmt_navy}{HTML}{000080}

\begin{acronym}[UMLX]
    \acro{DRY}{Don't Repeat Yourself}
    \acro{API}{Application Programming Interface}
    \acro{UML}{Unified Modeling Language}
    \acro{IR}{Information Retrieval}
    \acro{BDE}{Binary Direction Enhancement}
    \acro{FIMO}{finite iterable mathematical object}
    \acro{IIMO}{infinite iterable mathematical object}
    \acro{HD}{Hamming Distance}
    \acro{DCG}{discounted cumulative gain}
    \acro{MAP}{mean average precision}
    \acro{GMAP}{geometric mean of (per-topic) average precision}
    \acro{LDA}{Latent Dirichlet Allocation}
    \acro{LSA}{Latent Semantic Analysis}
    \acro{LSI}{Latent Semantic Indexing}
    \acro{NLP}{Natural Language Processing}
    \acro{ML}{machine learning}
    \acro{SVD}{singular value decomposition}
    \acro{TF}{term frequency}
    \acro{IDF}{inverse term frequency}
    \acro{CLIR}{Cross Language Information Retrieval}
    \acro{TMT}{Topic Model Translation}
    \acro{SMT}{statistical machine translation}
    \acro{NLTK}{Natural Language Toolkit}
    \acro{NLP}{Natural Language Processing}
    \acro{CLTM}{Cross-language topic modeling}
    \acro{NDCG}{normalized discounted cumulative gain}
    \acro{LLM}{Large Language Model}
    \acro{RR}{reciprocal rank}
    \acro{TP}[TP]{translation pair}
\end{acronym}

\begin{document}

\title{TMT: A Simple Way to Translate Topic Models Using Dictionaries}

\author{Felix Engl}
\email{felix.engl@uni-bamberg.de}
\orcid{0000-0003-1238-5449}
\affiliation{%
  \institution{University of Bamberg}
  \city{Bamberg}
  \country{Germany}
}

\author{Prof. Dr. Andreas Henrich}
\email{andreas.henrich@uni-bamberg.de}
\orcid{0000-0000-0000-0000}
\affiliation{%
  \institution{University of Bamberg}
  \city{Bamberg}
  \country{Germany}
}

\begin{abstract}
    The training of topic models for a multilingual environment is a challenging task, requiring the use of sophisticated algorithms, topic-aligned corpora, and manual evaluation. These difficulties are further exacerbated when the developer lacks knowledge of the target language or is working in an environment with limited data, where only small or unusable multilingual corpora are available.

    Considering these challenges, we introduce \ac{TMT}, a novel, robust and transparent technique designed to transfer topic models (e.g., \ac{LDA} based topic models) from one language to another, without the need for metadata, embeddings, or aligned corpora. \ac{TMT} enables the reuse of topic models across languages, making it especially suitable for scenarios where large corpora in the target language are unavailable or manual translation is infeasible. 
    Furthermore, we evaluate \ac{TMT} extensively using both quantitative and qualitative methods, demonstrating that it produces semantically coherent and consistent topic translations.
\end{abstract}

\acresetall



\maketitle

\section{Introduction}\label{sec:intro}%
The task of training topic models in a multilingual setting is challenging, especially when the data or expertise on the target language is limited. Existing methods for compensating these challenges usually rely on complex algorithms or large multilingual corpora, rendering the process demanding in various regards like available data or manual evaluation. In response, we introduce \ac{TMT}, a transparent, resource-efficient way to transfer topic models (e.g., \acs{LDA}-Models) from one language to another.

In short, \ac{TMT} is a robust algorithm that excels in scenarios where a given, ready-to-use monolingual topic model needs to cover multiple languages, or where only one language has a sufficient amount of training data and extensive parallel corpora are not available. This is done by, topic-wise, translating a monolingual topic model into another language via basic dictionaries and selecting the appropriate translations based on topical relevance, derived by voting techniques from expert search. We will also showcase a method to quantifying the consensus of the source and translated topic models by using established metrics like recall, precision and the \ac{NDCG}. The qualities of \ac{TMT} are particularly useful in digital humanities, enabling cross-linguistic comparisons of historical texts or training topic models for topics, where the target language has no sufficient data for training.

The rest of the publication is structured as follows: Section~2 reviews related works. Section~3 introduces \ac{TMT} and Section~4 describes the experimental setup. Section~5 discusses the experimental results and classifies the findings. The conclusions are stated in Section~6.%
\section{Background and Related Work}\label{sec:related}%
\ac{TMT} belongs to the field of \ac{CLTM}, but differs substantially from typical cross-lingual methods. Prior studies in this area, including \citet{Hao.2018}, primarily rely on large bilingual or multilingual corpora \cite{Sgaard.2015,Jagarlamudi.2010,BoydGraber.2012}, specialized datasets with meta information \cite{Zhao.2006}, cross-lingual topic-linked corpora \cite{Yang.2019}, or modifications to the underlying topic modeling algorithms \cite{Andrzejewski.2009,Dennis.1991,Bianchi.2020}. More recent approaches further integrate document relations \cite{Wang.2020,Terragni.2020}, word embeddings \cite{DeboraNozza.2016,Li.2016,Zhao.2017,Dieng.2020}, or pre-trained multilingual models such as BERT \cite{Devlin.2019,Vo.2022}.

In contrast, \ac{TMT} requires only a machine-readable dictionary that maps words between languages, and does not rely on additional metadata, contextualized datasets, or modifications to the topic modeling algorithm itself. This distinguishes \ac{TMT} as a more lightweight approach.

The approach most closely related to \ac{TMT} is described by \citet{Preiss.2012}, who uses a machine translation service to translate a topic model for evaluation purposes. However, this method is not described in detail and does not constitute a core contribution of that work.
\section{Methodology}\label{sec:method}%
\begin{figure*}
\centering
\includegraphics[
width=0.90\textheight,
height=0.90\textwidth,
viewport=0 0 1500 1200, 
clip,
angle=90,
keepaspectratio
]{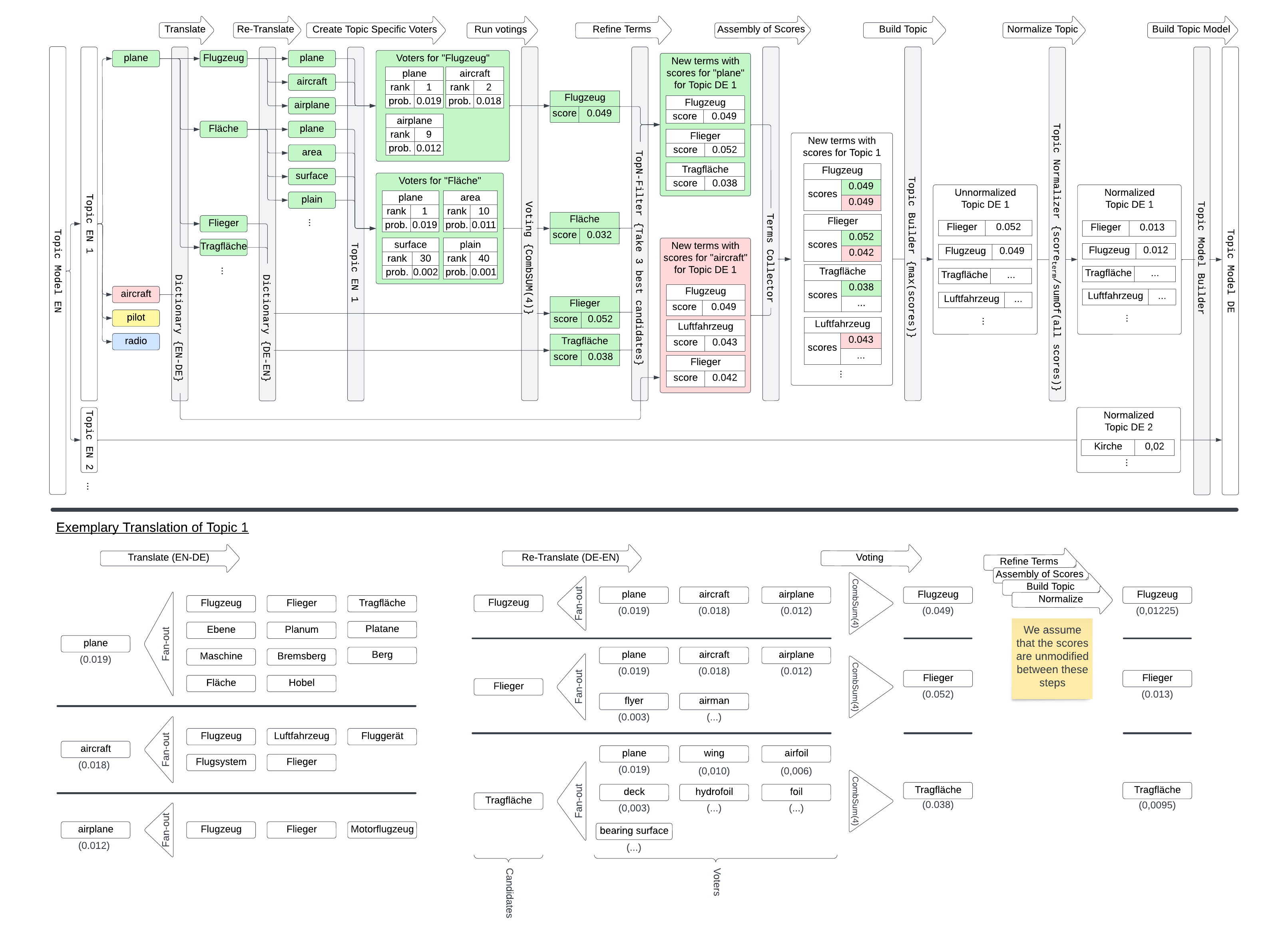}
\caption{An example translation of an English LDA topic model into German using \ac{TMT}. The translated topic 1 is about aviation. The curly brackets indicate the algorithms and models used. The second half of the figure shows the translation and re-translation process (including fan-out) as well as the voting and normalization step for three candidate terms in more detail.}
\Description{Exemplary translation of an English topic model for topic about aviation, the top four words are plane, aircraft, pilot, and radio. The process describes the translation of plane, aircraft , the steps are described in detail in the section Methodology - Abstract Algorithm.
1. Translate: plane is translated to Flugzeug, Fläche, Flieger, Tragfläche. These are the candidates.
2. Re-Translate
a. Flugzeug is re-translated to: plane, aircraft, and airplane.
b. Flieger is re-translated to: plane, area, surface, and plain.
3. Create Topic-Specific Voters
a. Flugzeug
- plane: rank 1, prob. 0.019
- aircraft: rank 2, rob. 0.018
- airplane: rank 9, rob. 0.012
b. Fläche
- plane: rank 1, prob. 0.019
- area: rank 10, prob. 0.011
- surface: rank 30, prob. 0.002
- plain: rank 40, prob. 0.001
4. Run Votings: CombSUM(4), sums up the top four voter scores for a candidate
a. Flugzeug: 0.049
b. Fläche: 0.032
c. Flieger: 0.052
d. Tragfläche: 0.038
5. Refine Terms: Take three best candidates for a English term
a. plane
- Flugzeug: 0.049
- Flieger: 0.052
- Tragfläche: 0.038
b. aircraft (steps 1 to 5 were omitted for this term)
- Flugzeug: 0.049
- Luftfahrzeug:  0.043
- Flieger: 0.042
6. Assembly of Scores
- Flugzeug: 0.049, 0.049
- Flieger: 0.052, 0.042
7. Build Topic: max(scores), results in an unnormalized, German topic 1
a. Flieger: 0.052
b. Flugzeug: 0.049
8. Normalize Topic: score(term)/sumOf(all scores)
a. Flieger: 0.013
b. Flugzeug: 0.012
9. Build Topic Model: Combine the translated topics to a topic model

The second part of the image shows the Translate, Re-Translate and Voting process in a bigger scale with more terms, to emphasize the fan-out effects. 
For example: The word plane translates to 11 German words. Four are related to aviation, two to geometry, two to construction, and three to mining.
}
\label{fig:voting}
\end{figure*}

This section explains \ac{TMT} by means of a step-by-step example (see~\cref{fig:voting}), followed by an introduction to the used terminology and voting models, as well as a detailed description of the \ac{TMT} algorithm.%
\subsection{Abstract Algorithm} \label{ssec:desc}%
The following text explains \ac{TMT} by means of an example translation of a single topic in \cref{fig:voting} using the algorithm, abstracted in nine steps.%
\paragraph{Translate}%
Step one uses a bidirectional dictionary to translate each word in a topic from language~A to language~B. Because words can have several translations between languages, this step usually results in a large and noisy set of candidate terms, with some of them not being related to the currently translated topic. This effect, hereafter referred to as \textit{fan-out}, is one of the biggest challenges of \ac{TMT}. We will describe the occurrence of this effect and the necessary countermeasures in the following (see~\cref{fig:voting}).%
\paragraph{Re-Translate}%
The same dictionary is then used in reverse to generate the re-translations, which may or may not reflect the original topic due to further fan-out. These re-translations from language~B to language~A are then used as voters for the language~B candidate terms in the following steps (see~\cref{fig:voting}, second half).%
\paragraph{Create Topic Specific Voters}%
The third step associates each re-translation to the membership probability for the target topic and the respective rank (positions in the topic's word list, sorted by score). As already mentioned, at this point the fan-out occurred twice. Resulting in multiple voters associated to a single candidate.%
\paragraph{Run Votings}\label{p:voting}%
To aggregate these membership probabilities in a meaningful way \ac{TMT} uses voting models to create so-called aggregated scores. In order to prevent a so-called topic degradation (loss of semantic coherence) due to the fan-out, it is important to select a voting model capable to properly aggregate the probabilities of the original topic model.

For example, the German translations of the word ``plane'' (see \cref{fig:voting}, second half) span multiple semantic domains, such as \textbf{aviation} ``Flugzeug/Fluggerät'' (``airplane/aircraft'') and \textbf{geography} ``Fläche/Ebene'' (``area/plain''). Naively assigning the original probability of ``plane'' to all translations across all topics would degrade the translated topics, by assigning high probabilities to semantically irrelevant words. (A detailed description of voting models is provided in \cref{ssec:voting}.)%

\paragraph{Refine Terms}%
As an optional fifth step, only the top $n$ candidates are kept ($n=3$ in \cref{ssec:voting}) for each translated word, in order to keep the translation concise by removing unsuitable candidates.%
\paragraph{Assembly of Scores}%
Due to fan-out multiple terms in the given topic can lead to the same translation. Therefore we collect all candidates of the same word in language~B and the corresponding aggregated scores. In our example ``Flieger'' is a translation of both ``plane'' and ``aircraft'', so all scores associated with the term ``Flieger'' have to be assembled. Depending on the voting model used, the assembled scores may be different for the same translation.%
\paragraph{Build Topic}%
This step merges the assembled scores from the previous step for each translation into a single value to create a distribution like representation. This can be done by simple aggregation methods, like $\text{max}$ (see~\cref{fig:voting}) or applying a voting model\footnote{In this publication we limit the aggregation to $\text{max}$. The application of more advanced techniques is subject for future research.}.%
\paragraph{Normalize Topic}%
The resulting distribution like representation is then normalized to achieve values similar to real probabilities.%
\paragraph{Build Topic Model}%
Applying these steps to each topic yields the final topic model in language~B.%
\subsection{Terminology}\label{ssec:term}
In order to describe \ac{TMT} as intuitively as possible, the notation and description of the algorithm is kept similar to the notation used in ``Latent Dirichlet Allocation'' by Blei et al. \cite{Blei.2003}. Thus, we define the following variables: 
\begin{enumerate}
    \item Let $L_X = \{\text{All words in language }X\}$.
    \item Let $\mathscr{D}^{X}_{Y}(w_X)$ be a primitive dictionary, mapping a single word $w$ from language~$X$ to multiple, or zero, words in language~$Y$.
    \item Let $T_X$ be a topic model for language~$X$.
    \item Let $K(T_X)$ be the number of topics for $T_X$.
    \item Let $V_X \subseteq L_X$ be the vocabulary of $T_X$.
    \item Let $\varphi^{X}_{k}$ be the word distribution for topic $k$ in $K(T_X)$.
    \item Let $\varphi^{X}_{k}(w)$ be the probability of word $w$ in $\varphi^{X}_{k}$.
    \item Let $\theta^{X}_{i}$ be the topic distribution of document $d_i$ with model $T_X$
    \item Let $\theta^{X}_{i}(k)$ be the probability of topic $k$ in $\theta^{X}_{i}$.
\end{enumerate}

\subsection{Algorithm}\label{ssec:alg}
Before describing the actual algorithm used by \ac{TMT} to translate each topic of a source model \TopicModelA into a target model \TopicModelB, we have to define three additional, manual optimization parameters:
\begin{enumerate} \item \Threshold filters out words in \VocabularyA with topic probabilities below the provided minimum threshold to reduce noise (see~\cref{alg:tmt:translate}~\cref{alg:tmt:translate:voters:filter}).
\item $n$ establishes a limit for the $n$-fitting translations produced by each translated entry of \TopicA (see \cref{alg:tmt:translate}~\cref{alg:tmt:translate:taken}).
\item \Epsilon is a fallback probability for words in \VocabularyB that were not translated for all topics (see \cref{alg:tmt:main}~\cref{alt:tmt:main:epsilon}); if not set explicitly, \Epsilon defaults to a value slightly lower than the smallest probability in the original and translated model.
\end{enumerate}%
\Cref{alg:tmt:main} applies \Translate to each topic, ensuring a parallelizable and self-contained translation process (see~\cref{alg:tmt:main}~\cref{alg:tmt:main:trans:loop}). After the  translation, \TopicModelB is assembled and \Epsilon is set. \FitToVocabuarly then ensures, that any missing words from \VocabularyB are added to \TopicB with the probability \Epsilon. Finally \TopicModelB is normalized to approximate a proper, topic model like, probability distribution.%

\begin{algorithm}
\caption{The main method of \acs{TMT} to translate a topic model from $L_A$ to $L_B$.}\label{alg:tmt:main}
\DontPrintSemicolon
\SetKwData{VocabularyCollection}{vocabularyCollection}
\SetKwData{Word}{word}
\SetKwData{VocabularyCollection}{vocabularyCollection}
\SetKwData{Tempt}{tmp}
\Fun{\TranslateTopicModel(\TopicModelA, \VocabularyA, \DictAB, \DictBA, \Epsilon\Gets\NULL, \Threshold\Gets\NULL, \NBest\Gets$\infty$)}{
    \KwResult{A translated topic model \TopicModelB and its vocabulary \VocabularyB}
    \BlankLine
    \tcp{Translate the Topic Model one topic at a time}
    \Val\TopicModelB\OfType\List[\Map{\String}{\Double}]\Gets\EmptyList\label{alg:tmt:main:trans:start}\;
    \For(\tcp*[f]{Loop over the topics}\label{alg:tmt:main:trans:loop}){$k$\OpIn$0$\OpTo$K(T_A)-1$}{
        \Val\TransTopicB\Gets\Translate{\TopicA, \VocabularyA, \DictAB, \DictBA, \Threshold,  \NBest}\label{alg:tmt:main:trans:pre}\;
        \TopicModelB\CallMember\Add{\TransTopicB}\label{alg:tmt:main:trans:end}\;
    }
    \BlankLine
    \tcp{Build the vocabulary}
    \Val\VocabularyCollection\OfType\HashSet[\String]\Gets\EmptyHashSet\label{alg:tmt:main:voc:start}\;
    \For{\TransTopicB\OpIn\TopicModelB}{
        \VocabularyCollection\CallMember\AddAll{\TransTopicB\AccessField\Keys}\;
    }

    \tcp{If \Epsilon is not set, use the smallest probability for the topic model.}
    \If{\label{alt:tmt:main:epsilon}\Epsilon $=$ \NULL}{
        \Temp\Gets\MinOf{\MinProbabilityIn{\TopicModelA}, \MinProbabilityIn{\TopicModelB}}\; \label{alg:tmt:main:smallest}
        \Epsilon\Gets\Temp $-$ \MachineEpsilon
    }
    \Val\VocabularyB\Gets\ConvertToList{\VocabularyCollection}\label{alg:tmt:main:voc:end}\;
    \tcp{Sets the words missing in each topic to \Epsilon and normalizes the vocabulary for each topic}
    \TopicModelB\CallMember\FitToVocabuarly{\VocabularyB, \Epsilon}\;\label{alg:tmt:main:repair}
    \TopicModelB\CallMember\Normalize{ };
    
    \KwRet (\TopicModelB, \VocabularyB)\;
}%
\end{algorithm}%

\cref{alg:tmt:translate} translates a single \TopicA of \TopicModelA to \TopicB with the aggregated scores (see~\cref{alg:tmt:main}~\cref{alg:tmt:main:trans:pre}). At first it collects all candidates an the respective voters for all entries in \VocabularyA, surpassing the threshold \Threshold (see~\cref{alg:tmt:translate} lines \ref{alg:tmt:translate:voters:filter}~-~\ref{alg:tmt:translate:voters:end}). Next, the translation process uses the voting model \ScoreModel to calculate a score for each candidate (see~\cref{alg:tmt:translate}~\cref{alg:tmt:translate:score}). \ScoreModel is a function that computes the aggregated score of a candidate by combining the scores from all voters, where higher values indicate stronger association with \TopicA. Finally, the algorithm selects the top $n$ candidates and adds them to the translated topic \TopicB using \AddAllHandleSameWord, ensuring that each candidate is added only once.%
\begin{algorithm}%
\DontPrintSemicolon%
\SetKwData{Probabilities}{probability}%
\SetKwData{X}{origin\textsubscript{A}}%
\SetKwData{Y}{trans\textsubscript{B}}%
\SetKwData{W}{reTrans\textsubscript{A}}%
\SetKwData{Rank}{rank}%
\SetKwData{Entry}{entry}%
\SetKwData{Translation}{trans\textsubscript{B}}%
\SetKwData{Voters}{voters}%
\SetKwData{Results}{results}%
\SetKwData{Y}{trans\textsubscript{B}}%
\caption{A function to translate a single topic from one language to another.}\label{alg:tmt:translate}%
\Fun{\Translate{\TopicA, \VocabularyA, \DictAB, \DictBA, \Threshold, \NBest}}{
    \KwResult{A translated topic \TopicB}

    \TypeAlias{\Voter}{\Tuple[\Rank\OfType\RankT, \Probabilities\OfType\ProbabilityT]}\;
    \TypeAlias{\Elected}{\Tuple[\Y\OfType\String, \VoteScore\OfType\Double]}\;
    \Var\TopicB\OfType\Map{\String}{\Double}\Gets\EmptyMap\;
    \For{\X\OpIn\VocabularyA}{
        \tcp{Ignore the word if its prob.\ is below the threshold \Threshold}
        \If{\TopicA\OpInvoke[\X] $\leq$ \Threshold\label{alg:tmt:translate:voters:filter}}{
            \Continue\;
        }
        \tcp{A mapping between candidates \Y and the voters from \TopicA}
        \Var\PartialTopics\OfType\Map{\String}{\List[\Voter]}\Gets\EmptyMap\label{alg:tmt:translate:voters:start}\;
        \For{\Y\OpIn\DictAB\OpInvoke[\X]}{
            \For{\W\OpIn\DictBA\OpInvoke[\Y]}{
                \Var\Probabilities\OfType\ProbabilityT\Gets\TopicA\OpInvoke[\W]\;
                \Var\Rank\OfType\RankT\Gets\TopicA\CallMember\GetRank{\W}\label{alg:tmt:translate:voters:rank}\;
                \PartialTopics\CallMember\GetOrCreate{\Y}\CallMember\Add{\Voter*[\Rank, \Probabilities]}\label{alg:tmt:translate:voters:end}\;
            }
        }
        \tcp{Calculate the scores for the candidates by the associated voters with a voting model.}
        \Var\ElectedCandidates\OfType\List[\Elected]\Gets\EmptyList\;
        \For{\Entry\OpIn\PartialTopics}{
            \Var(\Translation, \Voters)\Deconstruct\Entry\;
            \tcp{Calculates an aggregated score from all voters for a word.}
            \Var\VoteScore\Gets\ScoreModel\CallMember\Calculate{{\Voters}}\label{alg:tmt:translate:score}\;
            \ElectedCandidates\CallMember\Add{\Elected*[\Translation, \VoteScore]}\;
        }
        
        \tcp{Sort by the aggregated scores and take the best \NBest elements.}
        \Var\BestNCandidates\Gets\ElectedCandidates\CallMember\SortDescending{\Elected\AccessByReference\Second}\CallMember\Take{\NBest}\label{alg:tmt:translate:taken}\;
        \tcp{Duplicates are handled by \AddAllHandleSameWord}
        \TopicB\CallMember\AddAllHandleSameWord{\BestNCandidates}\label{alg:tmt:translate:add}\;
    }
    \KwRet\TopicB\;
}%
\end{algorithm}%
\subsection{Voting}\label{ssec:voting} 
As explained in \cref{ssec:desc}, voting models \cite{Balog.2012,Henrich.2021} are an integral part of \ac{TMT}. Formally, a voting model assigns scores to a set of classes $C$ (candidates) based on a query $q$, using a set of voters $R(q)$. Each voter is usually associated with one or more candidates in $C$. And the generalized scoring function of a voting model is $\Score_{<\text{name}>}(c, q, \cdot)$ calculating the score of a candidate $c \in C$ for a query $q$ and other voting-model-specific parameters.

In \ac{TMT}, the words from language~B are the candidates $C$, query $q$ is topic $k$, and the re-translations in language~A represent the voters $R(q)$. \Cref{tab:voting} outlines the voting models used to assess \ac{TMT}, including those from \cite{Balog.2012,Henrich.2021} as well as our own variants such as $\text{CombGSUM TOP}_n$, $\text{CombNOR TOP}_n$, and $\text{CombRR PEN TOP}_n$.

\begin{table*}
    \centering%
    \caption{Summary of the used voting models. For clarity, algorithm names are omitted after $\Score$ and indicated via suffixes (e.g., $\Score_{\text{Votes}}$). Variants with $n = \infty$ are excluded, and models using all voters omit the $\text{TOP}_n$ postfix.}
    \label{tab:voting}%
    \small%
    \begin{tabular}{lll}
        \toprule
        Name & Description & Formula \\
        \midrule
        $\text{Votes}$ & \makecell[l]{%
        Takes the number of found entries\\%
        for every candidate as the score.%
        } & 
        $\Score(c, q) = |\{a | a \in R(q) \wedge a \in c\}|$
        \\\hline
         
        $\text{CombSUM TOP}_n$ & \makecell[l]{%
        Sums up the scores from the $n \in [1, \infty]$\\%
        best entries for $c$ in $R(q)$.%
        } &
        $\Score(c, q, n)=\sum_{a \in R(q) \wedge a \in c \wedge \Rank(a, q, c) \leq n} \score(a, q)$
        \\\hline

        $\text{CombGSUM TOP}_n$ & \makecell[l]{%
        Calculates the geometric mean (\GeometricMean)\\%
        from the $n \in [1, \infty]$ best entries for\\%
        $c$ in $R(q)$.%
        } & 
        $\Score(c, q, n)=\GeometricMean(\set{\score(a, q) | a \in R(q) \wedge a \in c \wedge \Rank(a, q, c) \leq n})$
        \\\hline
         
        $\text{RR}^x\;\text{TOP}_n$ & \makecell[l]{%
        Sums up the reciprocal ranks from the\\%
        $n \in [1, \infty]$ best entries for $c \in R(q)$.\\%
        Exponent $x$ is used to adjust the shape\\%
        of the graph, its default value is 1.%
        } & 
        $\Score(c, q, n, x) = \sum_{a\in R(q) \wedge a \in c \wedge \Rank(a, q, c) \leq n} \left(\dfrac{1}{\Rank(a, q)}\right)^x$
        \\\hline

        $\text{CombSUM RR}^x\;\text{TOP}_n$ & \makecell[l]{%
        Combines $\text{CombSUM TOP}_n$ and\\%
        $\text{RR}^x\;\text{TOP}_n$ in order to utilize the\\%
        reciprocal rank as a penalty.%
        } & 
        $\Score(c, q, n, x) = \sum_{a \in R(q) \wedge a \in c \wedge \Rank(a, q, c) \leq n} \score(a, q) * \left(\dfrac{1}{\Rank(a, q)}\right)^x$
        \\\hline

        $\text{CombAVG TOP}_n$ & \makecell[l]{%
        Takes the average of the scores from the\\%
        $n \in [1, \infty]$ best entries for $c$ in $R(q)$.%
        } & 
        $\Score(c, q, n)=\overline{\set{\score(a, q) | a \in R(q) \wedge a \in c \wedge \Rank(a, q, c) \leq n}}$
        \\\hline

        $\text{CombNOR TOP}_n$ & \makecell[l]{%
        $\text{CombSUM TOP}_n$ where the score is\\%
        normalized by $\text{CombAVG TOP}_n$.%
        } & 
        $\Score(c, q, n) = \dfrac{\Score_{CombSUM\;TOP}(c, q, n) + \Score_{CombAVG\;TOP}(c, q, n)}{\Min(\Score_{Votes}(c, q), n) + 1}$
        \\\hline

        $\text{CombRR PEN TOP}_n$ & \makecell[l]{%
        Penalized combination of $\text{RR}^x\;\text{TOP}_n$\\%
        and $\text{CombAVG TOP}_n$. $\epsilon$ denotes the\\%
        smallest possible score for an entry $R(q)$.%
        } & 
        $\!\begin{aligned}
            \PComSUM(c, q, n) &= \begin{cases}
            \epsilon & \text{if } \Score_{Votes}(c, q) = 0 \\
            \frac{\Score_{CombSUM\;TOP}(c, q, n)}{\Min(\Score_{Votes}(c, q), n)} & \text{if } \Score_{Votes}(c, q) > 0
            \end{cases} \\
            \Score(c, q, n) &= \Score_{RR\;TOP}(c, q, 1, 1) + \PComSUM(c, q, n)
        \end{aligned}$
        \\\hline

        $\text{CombGNOR TOP}_n$ & \makecell[l]{%
        $\text{CombSUM TOP}_n$ where the score is\\%
        normalized by $\text{CombGSUM TOP}_n$.%
        } &  
        $\Score(c, q, n) = \dfrac{ \Score_{CombSUM\;TOP}(c, q, n) + \Score_{CombGSUM\;TOP}(c, q, n)}{\Min(\Score_{Votes}(c, q), n) + 1}$\\
        \bottomrule
    \end{tabular}
\end{table*}

\section{Experimental Setup}\label{sec:setup}%
To evaluate topic models translated with \ac{TMT}, a translation is considered successful if it preserves the model's relevance, coherence, and consistency across the source and target languages. Initially, standard coherence metrics (C\textsubscript{v}, u\textsubscript{mass}, c\textsubscript{w2v}) \cite{roedel.2015} were applied, but proved inadequate. Due to \ac{TMT}'s fan-out effect, even poor translations tend to achieve artificially high coherence scores, as the number of high-probability words increases regardless of true topic relevance. Moreover, since the translated topic models also lose their corpus-specific statistical properties, being represented solely by relative word probabilities derived from the voting process, this further distorts coherence-based evaluation. Heavily limiting the utility of conventional coherence metrics for evaluating \ac{TMT} by comparing it with other cross lingual methods, as \ac{TMT} translated topic models have unnaturally high coherence values.

As a result, we shifted the focus to topic consistency as the primary evaluation criterion. To measure the consistency, we trained an English topic model and translated it to German. Using aligned Wikipedia articles, we compared the inferred topic distributions from the original ``ideal'' English model to those of the translated topic models. Consistency is assessed using recall, precision, and \ac{NDCG} \cite{jarvelin.2002}.\footnote{Alternative metrics like Kendall Tau or Rank Biased Overlap (RBO) produced similar results.} These metrics enable comparison of the translations, as the inferred topic distribution $\theta_i$ for each document $i$ can be ranked by ordering the topics ($k \in K(T)$) according to their probability $\theta_i(k)$. The similarity between the translated and original topic models can then be quantified by applying \ac{NDCG} or similar ranking-based measures to these topic orderings.%
\subsection{Baselines}
We evaluate two simple baselines. The first, ``$\text{Plain}$'', directly assigns the original word's topic probability to each of its translations, without considering voters. Since all translations receive the same score, the $\text{TOP}_n$ variants of $\text{Plain}$ select $n$ candidates at random. The second baseline uses DeepL\footnote{\url{https://www.deepl.com/} (last visited 06.04.2025)} to translate each topic’s unprocessed vocabulary word-wise via the DeepL Python API. Contextual information is provided by passing a comma-separated list of high-probability topic terms (above the 80th-quantile) as the context argument. To retrieve the unprocessed (unstemmed) terms from the stemmed topic model vocabulary, we select the most frequent token from the corpus corresponding to the processed term in the topic model. 

\subsection{Dataset}\label{ssec:dattop}%
We use a pre-processed subset of WikiCOMP-2014 DE-EN from linguatools.org \footnote{see \href{https://linguatools.org/tools/corpora/wikipedia-comparable-corpora/}{https://linguatools.org/tools/corpora/wikipedia-comparable-corpora/} (last visit: 07.04.2025)}. After pre-processing the corpus, the remaining 251,845 aligned Wikipedia articles (\ds) are separated into 250,845 articles for training (\dsTrain) and 1,000 for testing (\dsTest).

The pre-processing removes in a first step the HTML-tags, then it segments the articles according to Unicode Standard Annex \#29, followed by normalizing the the segments with ``charabia'' \cite{meilisearchcharabia.2024} (version 0.8.11), and removing stopwords defined by the \acs{NLTK}-Framework \cite{bird2009natural} (version 3.8.1). The final two steps apply a language specific snowball stemmer and filter all aligned article pairs where one element contains less or more than 50 to 1,000 tokens.%
\subsection{Dictionary}%
The bidirectional dictionary \Dict is a composite of nine distinct lexicographic sources (see~\cref{tab:setup:dicts}) and consists of 2,043,260 bidirectional \acl{TP}s. The pipeline for composing \Dict starts by extracting all bidirectional \acp{TP}. The extracted \acp{TP} are then processed and tokenized by the same pipeline as \ds~(see \cref{ssec:dattop}), to make sure that the terms of \Dict are compatible with the English topic model and the translated ones. As a last step we removed duplicates and phrases longer than two terms.%
\begin{table}%
    \centering%
    \small%
    \caption{Overview of all dictionaries used to construct the composite dictionary for the \ac{TMT} evaluation. \#TP (\acl{TP}) is the total number of extracted de-en and en-de translations after being processed by the pipeline, excluding phrases longer than two terms. All links where last visited on 06.04.2025.}\label{tab:setup:dicts}
    \begin{tabular}{ccl}%
    \toprule%
    Name & \#TP & Link\\%
    \midrule%
    DictCC & 1,126,615 & {\href{https://www.dict.cc/}{www.dict.cc/}}\\%
    dicts.info & 148,543 & {\href{https://www.dicts.info/}{www.dicts.info/}}\\%
    ding & 699,669 & {\href{http://dict.tu-chemnitz.de/}{dict.tu-chemnitz.de/}}\\%
    EuroVoc & 17,191 & {\href{https://eur-lex.europa.eu/homepage.html}{https://eur-lex.europa.eu/}}\\%
    freedict & 268,985 & {\href{https://freedict.org/de/}{freedict.org/de/}}\\%
    IATE & 1,113,578 & {\href{https://iate.europa.eu/}{iate.europa.eu/}}\\%
    MS Terminology & 50,394 & {\href{https://learn.microsoft.com/en-us/globalization/reference/microsoft-language-resources}{learn.microsoft.com/}}\\%
    Wiktionary & 400,719 & {\href{https://de.wiktionary.org/wiki/Wiktionary:Hauptseite}{de.wiktionary.org/}}\\%
    MUSE & 352,671 & {\href{https://github.com/facebookresearch/MUSE}{github.com/facebookresearch/MUSE}}\\%
    \bottomrule%
    \end{tabular}%
\end{table}%
\subsection{Topic Model}%
Using \dsTrain\ we trained an English topic model $T_{en}$ with tomotopy, using \ac{LDA} with Gibbs Sampling  \footnote{see~\href{https://bab2min.github.io/tomotopy/}{https://bab2min.github.io/tomotopy/} (last visited: 20.05.2025)} version 0.12.4. The parameters used to train the topic model are described in \cref{tab:model:parameters} and not explicitly stated parameters were used with their default framework values.%
\begin{table}%
\centering%
\caption{Parameters used to train the tomotopy topic model.}\label{tab:model:parameters}%
\begin{tabular}{cccc}
    \toprule
    topic count (k) & term-weight & alpha & rm\_top \\
    \midrule
    25 & ONE (unweighted) & 0.01 & 5 \\
    \bottomrule
    \addlinespace
    \toprule
    min\_cf & burn\_in & \#iterations & \\
    \midrule
    3 & 100 & 1000 &  \\
    \bottomrule
\end{tabular}
\end{table}%
\subsection{TMT translation behavior}%
In addition to experimenting with different voting models, we evaluated six configurations that define how \ac{TMT} performs word-level translation (see~\cref{tab:configs}). The parameter ``Keep word from origin'' controls, whether original English terms are retained in the translated topic model: ``Always'' keeps the original term with its probability, ``Never'' drops the original word, and ``If no translation'' retains the term from the original if there is no translation found. The last approach enables \ac{TMT} to keep proper names, and therefore highly specialized terms, without adding too much noise by keeping all words from the original language. 
We also examined the impact of domain-specific dictionaries. Of the nine dictionaries listed in \cref{tab:setup:dicts}, EuroVoc and IATE are tailored for legal and administrative texts in the European Union, while the other seven are general-purpose. To assess the influence of domain content, we created two composite dictionary variants: one excluding EuroVoc and IATE, and one including all nine dictionaries (see~\cref{tab:configs}), allowing us to study how domain-specific vocabulary affects the translation quality of \ac{TMT}.%
\begin{table}
    \centering
    \caption{Additional parameters to change the word-level translation of \ac{TMT}.}
    \label{tab:configs}%
    \begin{tabular}{ccc}
        \toprule
        \multirow{2}{*}{tID} & \multicolumn{2}{c}{Translation Parameter}  \\
        & Keep word from origin & Exempt dictionaries \\
        \midrule
        0 & Always & EuroVoc, IATE \\
        1 & Never & EuroVoc, IATE \\
        2 & If no translation & EuroVoc, IATE \\
        3 & Always & - \\
        4 & Never & - \\
        5 & If no translation & - \\
        \bottomrule
    \end{tabular}
\end{table}

\subsection{Translations}\label{ssec:transl}%
To conduct a quantitative and qualitative evaluation (see~\cref{sec:result}), we translated $T_{en}$ to German $T_{ger}$, using 21 voting models and four baseline models with six different configurations (see~\cref{tab:configs}), resulting in a total of 145 translated topic models (see~\cref{tab:eval:combinations}). The syntax we will use to refer to a specific model by id is [A-X][0-5], meaning that J0 refers to ``RR'' with ``always keep the original word'', and ``ignore the dictionaries EuroVoc and IATE''. The color-coding used in \cref{tab:eval:combinations} is used to make the interpretation of the graph depicted in \cref{fig:top3comp} easier. In general, models depending on the reciprocal rank are color coded blue, models derived from $\text{CombSUM}$ are green and models derived from  $\text{CombNOR}$ are represented by red. The baselines are black.%
\begin{table}[t]
    \centering%
    \caption{A matrix of all analyzed translations, denoted by an unique ID composed from a voting model ID (vID) and a translation parameter ID (tID, see \cref{tab:configs}). The asterisks~(*) marks baseline models. The voting model names in the left column were constructed using the notation described in \cref{tab:voting}. Column \textit{C} provides a color-coding of the voting models.}\label{tab:eval:combinations}%
    \begin{tabular}{ccccccccc}
         \toprule
             \multirow{2}{*}{Voting Model} & \multirow{2}{*}{vID} & \multicolumn{6}{c}{tID} & \multirow{2}{*}{C}    \\
                                                          &    & 0  & 1  & 2  & 3  & 4  & 5  &                                        \\
         \midrule
            $\text{Plain}$*                               & A  & \parbox{7pt}{A0} & \parbox{7pt}{A1} & \parbox{7pt}{A2} & \parbox{7pt}{A3} & \parbox{7pt}{A4} & \parbox{7pt}{A5} & \colorbox{Baseline}{\phantom{x}} \\
            $\text{Plain}\;\text{TOP}_{2}$*               & B  & \parbox{7pt}{B0} & \parbox{7pt}{B1} & \parbox{7pt}{B2} & \parbox{7pt}{B3} & \parbox{7pt}{B4} & \parbox{7pt}{B5} & \colorbox{Baseline}{\phantom{x}} \\
            $\text{Plain}\;\text{TOP}_{3}$*               & C  & \parbox{7pt}{C0} & \parbox{7pt}{C1} & \parbox{7pt}{C2} & \parbox{7pt}{C3} & \parbox{7pt}{C4} & \parbox{7pt}{C5} & \colorbox{Baseline}{\phantom{x}} \\
            $\text{DeepL}$*                               & DL & \multicolumn{6}{c}{DL}                                                                                          & \colorbox{Baseline}{\phantom{x}} \\\hline
            $\text{CombSUM}$                              & D  & \parbox{7pt}{D0} & \parbox{7pt}{D1} & \parbox{7pt}{D2} & \parbox{7pt}{D3} & \parbox{7pt}{D4} & \parbox{7pt}{D5} & \colorbox{CombSUM}{\phantom{x}} \\
            $\text{CombSUM}\;\text{TOP}_{2}$              & E  & \parbox{7pt}{E0} & \parbox{7pt}{E1} & \parbox{7pt}{E2} & \parbox{7pt}{E3} & \parbox{7pt}{E4} & \parbox{7pt}{E5} & \colorbox{CombSUM}{\phantom{x}} \\
            $\text{CombSUM}\;\text{TOP}_{3}$              & F  & \parbox{7pt}{F0} & \parbox{7pt}{F1} & \parbox{7pt}{F2} & \parbox{7pt}{F3} & \parbox{7pt}{F4} & \parbox{7pt}{F5} & \colorbox{CombSUM}{\phantom{x}} \\\hline
            $\text{CombSUM}\;\text{RR}^2$                 & G  & \parbox{7pt}{G0} & \parbox{7pt}{G1} & \parbox{7pt}{G2} & \parbox{7pt}{G3} & \parbox{7pt}{G4} & \parbox{7pt}{G5} & \colorbox{CombSUMRR}{\phantom{x}} \\
            $\text{CombSUM}\;\text{RR}^2\;\text{TOP}_{2}$ & H  & \parbox{7pt}{H0} & \parbox{7pt}{H1} & \parbox{7pt}{H2} & \parbox{7pt}{H3} & \parbox{7pt}{H4} & \parbox{7pt}{H5} & \colorbox{CombSUMRR}{\phantom{x}} \\
            $\text{CombSUM}\;\text{RR}^2\;\text{TOP}_{3}$ & I  & \parbox{7pt}{I0} & \parbox{7pt}{I1} & \parbox{7pt}{I2} & \parbox{7pt}{I3} & \parbox{7pt}{I4} & \parbox{7pt}{I5} & \colorbox{CombSUMRR}{\phantom{x}} \\\hline
            $\text{RR}$                                   & J  & \parbox{7pt}{J0} & \parbox{7pt}{J1} & \parbox{7pt}{J2} & \parbox{7pt}{J3} & \parbox{7pt}{J4} & \parbox{7pt}{J5} & \colorbox{RR}{\phantom{x}} \\
            $\text{RR}\;\text{TOP}_{2}$                   & K  & \parbox{7pt}{K0} & \parbox{7pt}{K1} & \parbox{7pt}{K2} & \parbox{7pt}{K3} & \parbox{7pt}{K4} & \parbox{7pt}{K5} & \colorbox{RR}{\phantom{x}} \\
            $\text{RR}\;\text{TOP}_{3}$                   & L  & \parbox{7pt}{L0} & \parbox{7pt}{L1} & \parbox{7pt}{L2} & \parbox{7pt}{L3} & \parbox{7pt}{L4} & \parbox{7pt}{L5} & \colorbox{RR}{\phantom{x}} \\\hline
            $\text{CombGSUM}$                             & M  & \parbox{7pt}{M0} & \parbox{7pt}{M1} & \parbox{7pt}{M2} & \parbox{7pt}{M3} & \parbox{7pt}{M4} & \parbox{7pt}{M5} & \colorbox{CombGSUM}{\phantom{x}} \\
            $\text{CombGSUM}\;\text{TOP}_{2}$             & N  & \parbox{7pt}{N0} & \parbox{7pt}{N1} & \parbox{7pt}{N2} & \parbox{7pt}{N3} & \parbox{7pt}{N4} & \parbox{7pt}{N5} & \colorbox{CombGSUM}{\phantom{x}} \\
            $\text{CombGSUM}\;\text{TOP}_{3}$             & O  & \parbox{7pt}{O0} & \parbox{7pt}{O1} & \parbox{7pt}{O2} & \parbox{7pt}{O3} & \parbox{7pt}{O4} & \parbox{7pt}{O5} & \colorbox{CombGSUM}{\phantom{x}} \\\hline
            $\text{CombNOR}$                              & P  & \parbox{7pt}{P0} & \parbox{7pt}{P1} & \parbox{7pt}{P2} & \parbox{7pt}{P3} & \parbox{7pt}{P4} & \parbox{7pt}{P5} & \colorbox{CombNOR}{\phantom{x}} \\
            $\text{CombNOR}\;\text{TOP}_{2}$              & Q  & \parbox{7pt}{Q0} & \parbox{7pt}{Q1} & \parbox{7pt}{Q2} & \parbox{7pt}{Q3} & \parbox{7pt}{Q4} & \parbox{7pt}{Q5} & \colorbox{CombNOR}{\phantom{x}} \\
            $\text{CombNOR}\;\text{TOP}_{3}$              & R  & \parbox{7pt}{R0} & \parbox{7pt}{R1} & \parbox{7pt}{R2} & \parbox{7pt}{R3} & \parbox{7pt}{R4} & \parbox{7pt}{R5} & \colorbox{CombNOR}{\phantom{x}} \\\hline
            $\text{CombGNOR}$                             & S  & \parbox{7pt}{S0} & \parbox{7pt}{S1} & \parbox{7pt}{S2} & \parbox{7pt}{S3} & \parbox{7pt}{S4} & \parbox{7pt}{S5} & \colorbox{CombGNOR}{\phantom{x}} \\
            $\text{CombGNOR}\;\text{TOP}_{2}$             & T  & \parbox{7pt}{T0} & \parbox{7pt}{T1} & \parbox{7pt}{T2} & \parbox{7pt}{T3} & \parbox{7pt}{T4} & \parbox{7pt}{T5} & \colorbox{CombGNOR}{\phantom{x}} \\
            $\text{CombGNOR}\;\text{TOP}_{3}$             & U  & \parbox{7pt}{U0} & \parbox{7pt}{U1} & \parbox{7pt}{U2} & \parbox{7pt}{U3} & \parbox{7pt}{U4} & \parbox{7pt}{U5} & \colorbox{CombGNOR}{\phantom{x}} \\\hline
            $\text{CombRR PEN}$                           & V  & \parbox{7pt}{V0} & \parbox{7pt}{V1} & \parbox{7pt}{V2} & \parbox{7pt}{V3} & \parbox{7pt}{V4} & \parbox{7pt}{V5} & \colorbox{CombRRPEN}{\phantom{x}} \\
            $\text{CombRR PEN}\;\text{TOP}_{2}$           & W  & \parbox{7pt}{W0} & \parbox{7pt}{W1} & \parbox{7pt}{W2} & \parbox{7pt}{W3} & \parbox{7pt}{W4} & \parbox{7pt}{W5} & \colorbox{CombRRPEN}{\phantom{x}} \\
            $\text{CombRR PEN}\;\text{TOP}_{3}$           & X  & \parbox{7pt}{X0} & \parbox{7pt}{X1} & \parbox{7pt}{X2} & \parbox{7pt}{X3} & \parbox{7pt}{X4} & \parbox{7pt}{X5} & \colorbox{CombRRPEN}{\phantom{x}} \\
        \bottomrule
    \end{tabular}%
\end{table}%

\section{Evaluation}\label{sec:result}%
We evaluate \ac{TMT} using quantitative and qualitative methods after translating the English topic model $T_{en}$ with the 145 configurations from \cref{tab:eval:combinations}. The quantitative evaluation uses the \ac{NDCG}-based approach from \cref{sec:setup}, whereas qualitative analysis manually compares selected translations.%
\subsection{Quantitative Evaluation}\label{ssec:quant}%
\begin{figure}[t]%
\centering%
\caption{Comparing the quality of the translations achieved with the different configurations. The highlighted values are presented in higher detail in \cref{tab:top3sampleext}. The names of the model IDs are color-coded as described in \cref{tab:eval:combinations}.}\label{fig:top3comp}%

\Description{
Combined line and stacked bar chart comparing 145 translated topic models. Each bar corresponds to one model and shows how many aligned documents have 3, 2, 1, or 0 matching top-3 topics between the original English topic model and its German translation. These values are color-coded and also indicated in a legend. The black line above the bars represents the average NDCG@3 score per model.
Models are sorted from left to right by their average NDCG@3 value. The first group includes score-based TMT models with the vocabulary configurations "never" and "if no translation" achieving the highest NDCG@3 scores (above 0.720). These models also yield the highest number of documents with full topic alignment.

The second group contains reciprocal-rank-based TMT models. Their NDCG@3 scores range between 0.720 and 0.695. The order of the models is mainly decided by the vocabulary configuration, starting with "never" and "if no translation", directly followed by "always".

The final group includes baseline models, such as direct word mapping and DeepL-based translations, with NDCG@3 scores below 0.516. The DeepL baseline ranks 139th out of 145 and exhibits significantly fewer documents with three matching topics.

Selected configurations of interest are marked in the graph: P5, G5, P3, M3, C5*, DL*, and B3* (in order of their rank).
}
\includegraphics[width=0.90\textheight, angle=90]{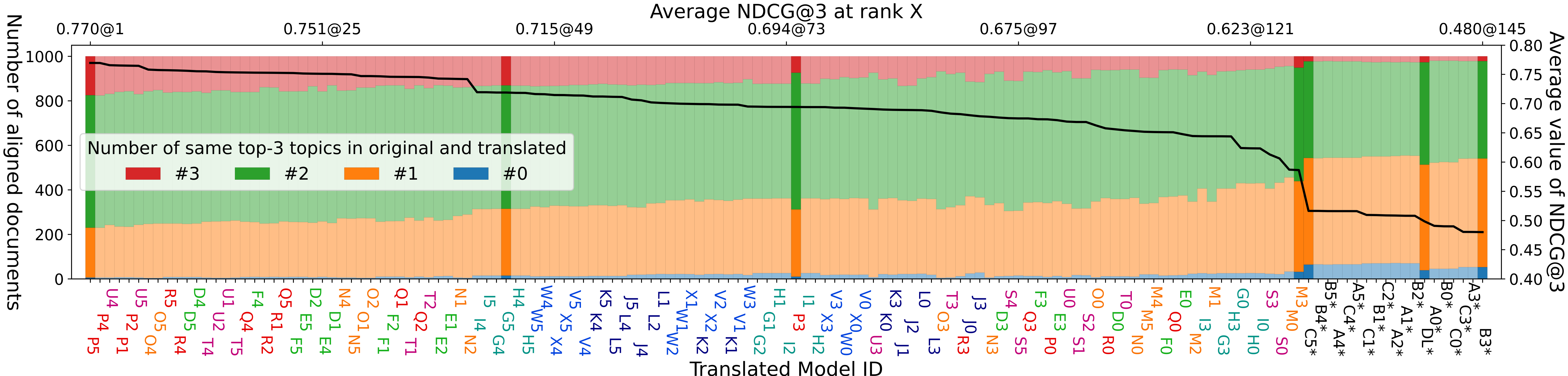}%
\end{figure}

The 145 German-translated models are compared against the original English model using $\mu$, the average NDCG@3 with rank-based weights (3, 2, 1). The objective is that, for the 1,000 aligned articles in \dsTest, the top three inferred topics in the German translations match, ideally retaining the order from the English original. In \cref{fig:top3comp} we represent these numbers with a line graph. Additionally, we count how many of the top three English topics appear in the German top three (ignoring order); three matches indicate the best case and zero the worst. This is also shown in \cref{fig:top3comp} as bars for ``Number of same top-3 topics in original and translated''. 

As illustrated in \cref{fig:top3comp}, baselines (marked with *) perform worse then any voting model based approach due to the inability to handle the fan-out. Because for baselines all translations inherit the original term's probability, even when semantically irrelevant. Looking at \cref{tab:top3sampleext}, even  the weakest voting-based translation (M3, $\mu=0.586$) outperforms the best baseline (C5, $\mu=0.516$) by $13.6\%$. The reason for DeepL's (DL*) under-performance lies within the nature of a \ac{LLM} and the constrains of the API and will be analyzed in \cref{ssec:qual}.

Voting models based on \ac{RR} (see \cref{tab:eval:combinations}, G-I, J-K, V-X) generally perform worse than score-based models due to \emph{hyperfocusing}, an effect where high-scoring voters are overemphasized while low-scoring ones are ignored, often amplifying dictionary errors. The sharp drop in average NDCG@3 between the best \ac{RR}-based model (I4) and the next score-based model highlights this issue (see \cref{fig:top3comp}). A more detailed example is provided in \cref{ssec:qual}. Limiting voters to the top-$n$ was only effective for geometric-mean-based models (see \cref{tab:eval:combinations}, M-O, S-U), where voters with low scores can similarly distort results. In contrast, arithmetic-mean-based models (see \cref{tab:eval:combinations}, D-F, P-R) benefit from considering all voters.

\begin{table}%
    \centering%
    \small%
    \caption{The statistical details about the selected NDCG@3 of the highlighted translations in \cref{fig:top3comp} over \dsTest. $\mu$ denotes the average NDCG@3 over all 1000 documents from \dsTest\ rounded to two digits.}\label{tab:top3sampleext}%
    \begin{tabular}{ccccccccccc}%
        \toprule
            & & P5 & G5 &  P3 & M3 & C5* & DL* & B3* \\
        \midrule
        \multicolumn{2}{c}{Rank}         & 1 & 44 & 74 & 126 & 127 & 139 & 145 \\
        \multicolumn{2}{c}{$\mu$}        & \parbox{13pt}{0.770} & \parbox{13pt}{0.719} & \parbox{13pt}{0.694} & \parbox{13pt}{0.586} & \parbox{13pt}{0.516} & \parbox{13pt}{0.498} & \parbox{13pt}{0.480} \\
        \bottomrule
        \addlinespace
        \toprule
        &  & P5 & G5 &  P3 & M3 & C5* & DL* & B3* \\
        \midrule
        \multirow{6}{30pt}{\makecell[c]{Aligned\\article\\count for\\NDCG@3}} & $1.00$ & 84 & 45 & 15 & 9 & 6 & 0 & 2\\
        & \parbox{30pt}{$]1.00, 0.75]$} & 552 & 481 & 423 & 276 & 183 & 143 & 124 \\
        & \parbox{30pt}{$]0.75, 0.50]$} & 288 & 350 & 429 & 393 & 365 & 366 & 364 \\
        & \parbox{30pt}{$]0.50, 0.25]$} & 70 & 109 & 123 & 290 & 382 & 452 & 457 \\
        & \parbox{30pt}{$]0.25, 0.00[$} & 59 & 87 & 95 & 231 & 310 & 389 & 377 \\
        & $0.00$ & 6 & 15 & 10 & 32 & 64 & 39 & 53\\
        \bottomrule
    \end{tabular}%
\end{table}%
Among the other evaluated parameters ``Keep word from origin'' and ``Exempt dictionaries'' (see \cref{tab:configs}), ``Keep word from origin'' has the strongest impact, even overpowering the voting model selection. Unconditionally keeping the original term (see \cref{tab:configs}, Config 0, 3) has a strong negative impact on the discerning power of the translations. Resulting in considerably worse translations, with the best being P3 at rank 74 (see \cref{fig:top3comp}). While voting models based on the \ac{RR} are able to mitigate the additional noise to some extend, non-\acs{RR} voting models proved to be incapable of maintaining the focus of the topics. On the other hand, keeping the original word when there is no translation or always dropping it only had a minor influence. But depending on the topic, keeping unknown terms could be beneficial for the translation process and will be examined in future research. 
In comparison the influence of ``Exempt dictionaries'' is minor, but in general \acp{TMT} prefer more complete dictionaries (3-5) over limited dictionaries (0-2) by properly minimizing the influence of the fan-out on the translation. 

We examined the six aligned articles from P5 with a \ac{NDCG} of $0.00$ (see \cref{tab:top3sampleext}) and 14 others with low scores across the ten best performing translated topic models. The primary challenge in evaluating \ac{TMT} using \ac{NDCG} was found to be content mismatch between aligned documents. In $50\%$ of the cases, articles focused on entirely different aspects of the same topic, causing the translated model to infer correct, but divergent, topics. One case involved a misalignment of two distinct individuals sharing the same name. Additional failures occurred due to inferring ``miscellaneous'' topics, such as name lists or acronym-heavy texts, lacking sufficient semantic context for reliable topic inference. Only one article was misranked due to fan-out; in all other cases, even low-scoring translations placed at least one reasonably accurate topic in the top three. Using \ac{TMT} as a method of judging the quality of aligned corpora or translations will be part of future research.%
\subsection{Qualitative Evaluation}\label{ssec:qual}%
As noted in \cref{ssec:quant}, the goal of \ac{TMT} is not to eliminate fan-out, but to leverage it in a controlled manner to enrich translated topic models with semantically relevant terms. This becomes evident when examining a translated topic focused on Scandinavian and Asian countries (see \cref{tab:eval:examplemodel}). Due to space constrains, we focus on term ranks 1-6 and 11-15. Terms with identical probabilities, due to fan-out, are grouped into single cells. The table should be read column-wise (top to bottom). Rows are used as reference only and do not imply semantic equivalence between terms.

\begin{table*}%
    \centering%
    \caption{Comparison table for a Scandinavian/Asian topic. The table shows the original $T_{en}$ topic model and selected translations from \cref{fig:top3comp}. Word probabilities are given in brackets. Phrases were excluded, as they do not affect the topic inference.}\label{tab:eval:examplemodel}%
    \small%
    \begin{tabular}{p{5pt}cccccccc}
        \toprule
        \phantom{x} & $T_{en}$ & P5 & G5 & P3 & M3 & C5* & DL* & B3*\\
         \midrule
        1 & \makecell[c]{swedish\\(0.01383)} & \makecell[c]{schwedisch\\schwedischsein\\(0.00354)} & \makecell[c]{schwed\\(0.18393)} & \makecell[c]{schwedischsein\\(0.00344)} & \makecell[c]{schwedischsein\\(0.00448)} & \makecell[c]{schwedisch\\schwedischsein\\(0.00473)} & \makecell[c]{schwed\\(0.01383)} & \makecell[c]{schwedischsein\\(0.00437)}\\
        2 & \makecell[c]{japanes\\(0.01194)} & \makecell[c]{japanerin\\japanisch\\(0.00305)} & \makecell[c]{schwedisch\\schwedischsein\\(0.17587)} & \makecell[c]{japanerin\\(0.00297)} & \makecell[c]{japanerin\\(0.00387)} & \makecell[c]{japanerin\\japanisch\\(0.00409)} & \makecell[c]{japan\\(0.01194)} & \makecell[c]{japanerin\\(0.00377)}\\
        3 & \makecell[c]{japan\\(0.01069)} & \makecell[c]{japanisi\\japanlack\\(0.00273)} & \makecell[c]{japan\\(0.05307)} & \makecell[c]{japanisi\\japanlack\\(0.00266)} & \makecell[c]{japanisi\\japanlack\\(0.00346)} & \makecell[c]{japanisi\\japanlack\\(0.00366)} & \makecell[c]{japan\\(0.01069)} & \makecell[c]{japanisi\\japanlack\\(0.00338)}\\
        4 & \makecell[c]{sweden\\(0.00973)} & \makecell[c]{norwegerin\\norwegischsprach\\norwegischsprech\\(0.00233)} & \makecell[c]{japanerin\\japanisch\\(0.03797)} & \makecell[c]{sweden\\(0.00242)} & \makecell[c]{sweden\\(0.00315)} & \makecell[c]{norwegerin\\norwegischsprach\\norwegischsprech\\(0.00312)} & \makecell[c]{schwed\\(0.00973)} & \makecell[c]{sweden\\(0.00307)}\\
        5 & \makecell[c]{norwegian\\(0.00912)} & \makecell[c]{danisch\\plundergeback\\plunderstuck\\(0.00212)} & \makecell[c]{kategori\\(0.02630)} & \makecell[c]{norwegerin\\norwegischsprach\\norwegischsprech\\(0.00227)} & \makecell[c]{norwegerin\\norwegischsprach\\norwegischsprech\\(0.00296)} & \makecell[c]{danisch\\plundergeback\\plunderstuck\\(0.00284)} & \makecell[c]{norweg\\(0.00912)} & \makecell[c]{norwegerin\\norwegischsprach\\norwegischsprech\\(0.00288)}\\
        6 & \makecell[c]{danish\\(0.00830)} & \makecell[c]{japan\\(0.00193)} & \makecell[c]{nippon\\(0.01511)} & \makecell[c]{plundergeback\\plunderstuck\\(0.00206)} & \makecell[c]{plundergeback\\plunderstuck\\(0.00269)} & \makecell[c]{medizinalrhabarb\\(0.00247)} & \makecell[c]{danisch\\(0.00830)} & \makecell[c]{plundergeback\\plunderstuck\\(0.00262)}\\
        \bottomrule
        \addlinespace
         \midrule
        11 & \makecell[c]{name\\(0.00591)} & \makecell[c]{benams\\verzeichnisnam\\(0.00151)} & \makecell[c]{norweg\\(0.00606)} & \makecell[c]{benams\\name\\verzeichnisnam\\(0.00147)} & \makecell[c]{stockholm\\stockholmerin\\(0.00139)} & \makecell[c]{stockholm\\stockholmerin\\(0.00146)} & \makecell[c]{nam\\(0.00591)} & \makecell[c]{stockholm\\stockholmerin\\(0.00135)}\\
        12 & \makecell[c]{denmark\\(0.00532)} & \makecell[c]{schwed\\(0.00148)} & \makecell[c]{norwegerin\\norwegischsprach\\norwegischsprech\\(0.00464)} & \makecell[c]{schwed\\(0.00143)} & \makecell[c]{tokiot\\(0.00130)} & \makecell[c]{tokio\\tokiot\\(0.00137)} & \makecell[c]{danemark\\(0.00532)} & \makecell[c]{tokiot\\(0.00126)}\\
        13 & \makecell[c]{stockholm\\(0.00428)} & \makecell[c]{nippon\\(0.00141)} & \makecell[c]{dan\\(0.00293)} & \makecell[c]{japanes\\(0.00133)} & \makecell[c]{kurrhahn\\putenfleisch\\trut\\trutenfleisch\\truthahnfleisch\\truthenn\\truthuhn\\(0.00129)} & \makecell[c]{kurr\\kurrhahn\\putenfleisch\\trut\\trutenfleisch\\truthahnfleisch\\truthenn\\truthuhn\\(0.00137)} & \makecell[c]{stockholm\\(0.00428)} & \makecell[c]{kurrhahn\\putenfleisch\\trut\\trutenfleisch\\truthahnfleisch\\truthenn\\truthuhn\\(0.00126)}\\
        14 & \makecell[c]{tokyo\\(0.00400)} & \makecell[c]{nihon\\(0.00139)} & \makecell[c]{plund\\(0.00293)} & \makecell[c]{norweg\\(0.00119)} & \makecell[c]{croatian\\krawot\\(0.00128)} & \makecell[c]{krawot\\(0.00135)} & \makecell[c]{tokio\\(0.00400)} & \makecell[c]{croatian\\krawot\\(0.00124)}\\
        15 & \makecell[c]{turkey\\(0.00399)} & \makecell[c]{nisei\\(0.00137)} & \makecell[c]{danisch\\plundergeback\\plunderstuck\\(0.00293)} & \makecell[c]{stockholm\\stockholmerin\\(0.00106)} & \makecell[c]{iceland\\islanderin\\islanderinn\\(0.00121)} & \makecell[c]{islanderin\\islanderinn\\(0.00128)} & \makecell[c]{turkei\\(0.00399)} & \makecell[c]{iceland\\islanderin\\islanderinn\\(0.00118)}\\
        \bottomrule
    \end{tabular}%
\end{table*}%
The DeepL baseline (DL*) in \cref{tab:eval:examplemodel} illustrates the key challenges when translating topic models using \acp{LLM}. A major limitation is that DeepL's API does not support alternative translations, resulting in direct lexical mapping that often duplicates terms, e.g. ``japan'' appears at both ranks 2 and 3 (see~\cref{tab:eval:examplemodel}), causing overfitting and reduced topic separability. The Google Translate API (checked on 21.04.2025) exhibits similar limitations. In contrast, \ac{TMT} benefits from controlled fan-out, as seen in the translation of ``japanese'' and ``Japan'' in \cref{tab:eval:japanese}. \ac{TMT} not only produces expected translations (e.g., demonyms), but also enriches the topic with relevant terms such as adjectives, loanwords (``nisei''), and colloquialisms. It also captures rare synonyms in the target language, such as ``Nihon'' and ``Nippon''. As observed with P5 (see~\cref{tab:eval:examplemodel}) these variants usually have different probabilities in the translated topic model, depending on their respective voters. A second issue with \acs{LLM}-based translations is the generation of multi-word phrases of varying lengths, which are incompatible with most topic modeling techniques and pre-processing pipelines on single token level. %
\begin{table}[b]%
\centering
\small
\caption{Translations from the composed dictionary used by \ac{TMT}. The unstemmed terms are written in {[\dots]}. A dagger (\textdagger) marks colloquial words.}\label{tab:eval:japanese}
\begin{tabular}{ccl}
\toprule
Original Term & Translated Term & Meaning \\
\midrule
\multirow{4}{*}[-11pt]{\makecell{japanes\\ {[japanese]}}} & \makecell{japan\\ {[Japaner]} \\ {[Japanerin]}} & \makecell[l]{A man/woman from Japan.} \\
& \makecell{japanisch\\ {[japanisch]}} & \makecell[l]{Adjective describing that some-\\thing is of Japanese origin.}\\
& \makecell{japanisi\\ {[japanisiert\textsuperscript{\textdagger}]}} & \makecell[l]{Describing that something is/was\\influenced by Japanese culture.}\\
& \makecell{nisei\\ {[Nisei]}} & \makecell[l]{Japanese term to describe\\ethnically Japanese second\\generation immigrants in\\North and South America.} \\\addlinespace
\makecell{japanes\\ {[japanese\textsuperscript{\textdagger}]}}  & \makecell{japanlack\\ {[Japanlack]}} & \makecell[l]{Describing lacquer used for\\japanese lacquerware.}\\\hline
\multirow{3}{*}[-10pt]{\makecell{japan\\ {[Japan]}} } & \makecell{japan\\ {[Japan]}} & Official German name for japan.\\
& \makecell{nihon\\ {[Nihon]}}  & \multirow{3}{*}{\makecell[l]{The name for Japan in Japanese\\when read aloud, sometimes used\\in German.}} \\
& \makecell{nippon\\ {[Nippon]}} & \\
\bottomrule
\end{tabular}
\end{table}%

As discussed in \cref{ssec:quant}, hyperfocusing is a major challenge for rank-based voting models. This effect occurs when voting models give too much weight to a few high-scoring candidates while ignoring others, often leading to overfitting or semantic drift. This is also reflected in what we call topic sharpness, which measures how quickly word probabilities drop by calculating the slope between two ranks in a topic model. In the provided data in \cref{tab:eval:examplemodel}, comparing the sharpness for the top five terms, $T_{en}$ has a moderate sharpness ($m_{\text{Top5}} = (0.00912 - 0.01383)/(1-5) = 0.0012$) and P5 is flat ($m_{\text{Top5}} = 0.0002$) too, but the \acs{RR}-based G5 shows a high topic sharpness ($m_{\text{Top5}} = 0.03649$). This steep slope suggests, that only a few terms represent the topic while the other terms only play a minor role when inferring the topics for a text. When we analyzed the topic sharpness for the top 100 terms for all translations shown in \cref{fig:top3comp}, we observed that well-performing translated models generally have flatter slopes than the original $T_{en}$, since translated topic models usually have a bigger vocabulary than the original due to fan-out. Suggesting that a similar or even flatter slope, than the originals, leads to better translations. Wether this topic sharpness can serve as a basis for a, corpus independent, quality metric for \ac{TMT} will be subject of future research.

Another challenge for \ac{TMT} is \emph{homonym noise}, because \ac{TMT} is context-ignorant it treats every term, even unrelated homonyms, as equally valid candidates for translation. For example in \cref{tab:eval:examplemodel}, the term ``danish'' in $T_{en}$ at rank 6 refers to the nation and nationality. However P5 translates it to ``Dänisch'' (demonym) and ``Plundergebäck'' (pastry) and assigns the same score to both of them, introducing unintended noise into the translated topic. 

Overall, the analysis demonstrates \ac{TMT}'s capability to translate topic models in a meaningful way. It also highlights the critical role of controlled fan-out and the voting model's capability to manage it. Additionally, we identified multiple challenges and opportunities such as hyperfocusing, homonym noise or topic sharpness for future research.%
\section{Conclusion}\label{sec:conclusion}%
In this paper we presented \ac{TMT}, a new approach for cross lingual topic modeling by translating pre-trained topic models. The evaluation of \ac{TMT} indicates, that it produces meaninful translations and outperforms direct approaches such as direct translations or using translation services. A central element for \ac{TMT} is the management of fan-out, by selecting a fitting voting model. Furthermore we found a relation between the topic sharpness and performance of the translated topic model. Challenges such as hyperfocusing and homonym noise remain. They will be subject to future research, including context awareness, topic sharpness as metric, and using \ac{TMT} for improving the quality of aligned corpora. The source code for \ac{TMT} is provided on GitHub at \url{https://github.com/FelixEngl/ptmt}. 

\clearpage

\section*{Generative AI Disclosure}

The following outlines the use of generative AI tools in the preparation of this publication:
\paragraph{Translation for experiments.}
DeepL's translation API was used to create a contextualized baseline for \ac{TMT} by translating the topics of an English topic model into German. The experiment using the DeepL API was run between the 12th and 14th December 2024.
\paragraph{No AI-generated research content.}
No generative AI was used to create new content for this publication in the form of text, images, tables, source code, data, or in any other undisclosed manner.
\paragraph{Writing assistance.}
"DeepL Write Pro" and "ChatGPT" (versions 4.0, 4.5, and 1o) were used as writing assistants. Their usage was limited to grammar correction, clarity improvement, and suggesting shorter or alternative phrasings for author-written text.

\bibliographystyle{ACM-Reference-Format}
\bibliography{references2}


\begin{thebibliography}{25}


\ifx \showCODEN    \undefined \def \showCODEN     #1{\unskip}     \fi
\ifx \showISBNx    \undefined \def \showISBNx     #1{\unskip}     \fi
\ifx \showISBNxiii \undefined \def \showISBNxiii  #1{\unskip}     \fi
\ifx \showISSN     \undefined \def \showISSN      #1{\unskip}     \fi
\ifx \showLCCN     \undefined \def \showLCCN      #1{\unskip}     \fi
\ifx \shownote     \undefined \def \shownote      #1{#1}          \fi
\ifx \showarticletitle \undefined \def \showarticletitle #1{#1}   \fi
\ifx \showURL      \undefined \def \showURL       {\relax}        \fi
\providecommand\bibfield[2]{#2}
\providecommand\bibinfo[2]{#2}
\providecommand\natexlab[1]{#1}
\providecommand\showeprint[2][]{arXiv:#2}

\bibitem[Andrzejewski et~al\mbox{.}(2009)]%
        {Andrzejewski.2009}
\bibfield{author}{\bibinfo{person}{David Andrzejewski}, \bibinfo{person}{Xiaojin Zhu}, {and} \bibinfo{person}{Mark Craven}.} \bibinfo{year}{2009}\natexlab{}.
\newblock \showarticletitle{Incorporating Domain Knowledge into Topic Modeling via Dirichlet Forest Priors}. In \bibinfo{booktitle}{\emph{ICML'09: Proceedings of the 26th Annual International Conference on Machine Learning}}, Vol.~\bibinfo{volume}{382}. \bibinfo{publisher}{{Association for Computing Machinery}}, \bibinfo{address}{New York, NY, USA}, \bibinfo{pages}{25--32}.
\newblock
\showISBNx{9781605585161}
\href{https://doi.org/10.1145/1553374.1553378}{doi:\nolinkurl{10.1145/1553374.1553378}}


\bibitem[Balog et~al\mbox{.}(2012)]%
        {Balog.2012}
\bibfield{author}{\bibinfo{person}{Krisztian Balog}, \bibinfo{person}{Yi Fang}, \bibinfo{person}{Maarten de Rijke}, \bibinfo{person}{Pavel Serdyukov}, {and} \bibinfo{person}{Luo Si}.} \bibinfo{year}{2012}\natexlab{}.
\newblock \showarticletitle{Expertise Retrieval}.
\newblock \bibinfo{journal}{\emph{Foundations and Trends® in Information Retrieval}} \bibinfo{volume}{6}, \bibinfo{number}{2–3} (\bibinfo{year}{2012}), \bibinfo{pages}{127--256}.
\newblock
\showISSN{1554-0669}
\href{https://doi.org/10.1561/1500000024}{doi:\nolinkurl{10.1561/1500000024}}


\bibitem[Bianchi et~al\mbox{.}(2021)]%
        {Bianchi.2020}
\bibfield{author}{\bibinfo{person}{Federico Bianchi}, \bibinfo{person}{Silvia Terragni}, \bibinfo{person}{Dirk Hovy}, \bibinfo{person}{Debora Nozza}, {and} \bibinfo{person}{Elisabetta Fersini}.} \bibinfo{year}{2021}\natexlab{}.
\newblock \showarticletitle{Cross-lingual Contextualized Topic Models with Zero-shot Learning}. In \bibinfo{booktitle}{\emph{Proceedings of the 16th Conference of the European Chapter of the Association for Computational Linguistics: Main Volume}}, \bibfield{editor}{\bibinfo{person}{Paola Merlo}, \bibinfo{person}{Jorg Tiedemann}, {and} \bibinfo{person}{Reut Tsarfaty}} (Eds.). \bibinfo{publisher}{Association for Computational Linguistics}, \bibinfo{address}{Online}, \bibinfo{pages}{1676--1683}.
\newblock
\href{https://doi.org/10.18653/v1/2021.eacl-main.143}{doi:\nolinkurl{10.18653/v1/2021.eacl-main.143}}


\bibitem[Bird et~al\mbox{.}(2009)]%
        {bird2009natural}
\bibfield{author}{\bibinfo{person}{Steven Bird}, \bibinfo{person}{Ewan Klein}, {and} \bibinfo{person}{Edward Loper}.} \bibinfo{year}{2009}\natexlab{}.
\newblock \bibinfo{booktitle}{\emph{Natural Language Processing with Python}}.
\newblock \bibinfo{publisher}{O'Reilly Media, Inc.}
\newblock
\showISBNx{9780596516499}


\bibitem[Blei et~al\mbox{.}(2003)]%
        {Blei.2003}
\bibfield{author}{\bibinfo{person}{David~M. Blei}, \bibinfo{person}{Andrew~Y. Ng}, {and} \bibinfo{person}{Michael~I. Jordan}.} \bibinfo{year}{2003}\natexlab{}.
\newblock \showarticletitle{Latent dirichlet allocation}.
\newblock \bibinfo{journal}{\emph{J. Mach. Learn. Res.}}  \bibinfo{volume}{3} (\bibinfo{date}{March} \bibinfo{year}{2003}), \bibinfo{pages}{993–1022}.
\newblock
\showISSN{1532-4435}


\bibitem[Boyd-Graber and Blei(2009)]%
        {BoydGraber.2012}
\bibfield{author}{\bibinfo{person}{Jordan Boyd-Graber} {and} \bibinfo{person}{David~M. Blei}.} \bibinfo{year}{2009}\natexlab{}.
\newblock \showarticletitle{Multilingual topic models for unaligned text}. In \bibinfo{booktitle}{\emph{Proceedings of the Twenty-Fifth Conference on Uncertainty in Artificial Intelligence}} (Montreal, Quebec, Canada) \emph{(\bibinfo{series}{UAI '09})}. \bibinfo{publisher}{AUAI Press}, \bibinfo{address}{Arlington, Virginia, USA}, \bibinfo{pages}{75–82}.
\newblock
\showISBNx{9780974903958}
\urldef\tempurl%
\url{https://dl.acm.org/doi/10.5555/1795114.1795124}
\showURL{%
\tempurl}


\bibitem[{Debora Nozza} et~al\mbox{.}(2016)]%
        {DeboraNozza.2016}
\bibfield{author}{\bibinfo{person}{{Debora Nozza}}, \bibinfo{person}{{Elisabetta Fersini.}}, {and} \bibinfo{person}{{Enza Messina.}}} \bibinfo{year}{2016}\natexlab{}.
\newblock \showarticletitle{Unsupervised Irony Detection: A Probabilistic Model with Word Embeddings}.
\newblock \bibinfo{journal}{\emph{2184-3228}} (\bibinfo{year}{2016}).
\newblock
\showISSN{2184-3228}
\href{https://doi.org/10.5220/0006052000680076}{doi:\nolinkurl{10.5220/0006052000680076}}


\bibitem[Dennis(1991)]%
        {Dennis.1991}
\bibfield{author}{\bibinfo{person}{Samuel~Y. Dennis}.} \bibinfo{year}{1991}\natexlab{}.
\newblock \showarticletitle{On the hyper-dirichlet type 1 and hyper-liouville distributions}.
\newblock \bibinfo{journal}{\emph{Communications in Statistics - Theory and Methods}} \bibinfo{volume}{20}, \bibinfo{number}{12} (\bibinfo{year}{1991}), \bibinfo{pages}{4069--4081}.
\newblock
\showISSN{0361-0926}
\href{https://doi.org/10.1080/03610929108830757}{doi:\nolinkurl{10.1080/03610929108830757}}


\bibitem[Devlin et~al\mbox{.}(2019)]%
        {Devlin.2019}
\bibfield{author}{\bibinfo{person}{Jacob Devlin}, \bibinfo{person}{Ming-Wei Chang}, \bibinfo{person}{Kenton Lee}, {and} \bibinfo{person}{Kristina Toutanova}.} \bibinfo{year}{2019}\natexlab{}.
\newblock \showarticletitle{BERT: Pre-training of Deep Bidirectional Transformers for Language Understanding}. In \bibinfo{booktitle}{\emph{NAACL'19: Proceedings of the 2019 Conference of the North American Chapter of the Association for Computational Linguistics: Human Language Technologies}}, \bibfield{editor}{\bibinfo{person}{Jill Burstein}, \bibinfo{person}{Christy Doran}, {and} \bibinfo{person}{Thamar Solorio}} (Eds.). \bibinfo{publisher}{{Association for Computational Linguistics}}, \bibinfo{address}{Minneapolis, Minnesota}, \bibinfo{pages}{4171--4186}.
\newblock
\href{https://doi.org/10.18653/v1/N19-1423}{doi:\nolinkurl{10.18653/v1/N19-1423}}


\bibitem[Dieng et~al\mbox{.}(2020)]%
        {Dieng.2020}
\bibfield{author}{\bibinfo{person}{Adji~B. Dieng}, \bibinfo{person}{Francisco J.~R. Ruiz}, {and} \bibinfo{person}{David~M. Blei}.} \bibinfo{year}{2020}\natexlab{}.
\newblock \showarticletitle{Topic Modeling in Embedding Spaces}.
\newblock \bibinfo{journal}{\emph{Transactions of the Association for Computational Linguistics}}  \bibinfo{volume}{8} (\bibinfo{date}{07} \bibinfo{year}{2020}), \bibinfo{pages}{439--453}.
\newblock
\showISSN{2307-387X}
\href{https://doi.org/10.1162/tacl_a_00325}{doi:\nolinkurl{10.1162/tacl_a_00325}}


\bibitem[Hao and Paul(2018)]%
        {Hao.2018}
\bibfield{author}{\bibinfo{person}{Shudong Hao} {and} \bibinfo{person}{Michael~J. Paul}.} \bibinfo{year}{2018}\natexlab{}.
\newblock \bibinfo{title}{An Empirical Study on Crosslingual Transfer in Probabilistic Topic Models}.
\newblock
\urldef\tempurl%
\url{http://arxiv.org/pdf/1810.05867v2}
\showURL{%
\tempurl}


\bibitem[Henrich and Wegmann(2021)]%
        {Henrich.2021}
\bibfield{author}{\bibinfo{person}{Andreas Henrich} {and} \bibinfo{person}{Markus Wegmann}.} \bibinfo{year}{2021}\natexlab{}.
\newblock \showarticletitle{Search and Evaluation Methods for Class Level Information Retrieval: Extended Use and Evaluation of Methods Applied in Expertise Retrieval}. In \bibinfo{booktitle}{\emph{SAC '21: Proceedings of the 36th Annual ACM Symposium on Applied Computing}}, \bibfield{editor}{\bibinfo{person}{{Association for Computational Linguistics}}} (Ed.). \bibinfo{publisher}{{Association for Computing Machinery}}, \bibinfo{address}{New York, NY, USA}, \bibinfo{pages}{681--684}.
\newblock
\showISBNx{9781450381048}
\href{https://doi.org/10.1145/3412841.3442092}{doi:\nolinkurl{10.1145/3412841.3442092}}


\bibitem[Jagarlamudi and Daum{\'e}(2010)]%
        {Jagarlamudi.2010}
\bibfield{author}{\bibinfo{person}{Jagadeesh Jagarlamudi} {and} \bibinfo{person}{Hal Daum{\'e}}.} \bibinfo{year}{2010}\natexlab{}.
\newblock \showarticletitle{Extracting Multilingual Topics from Unaligned Comparable Corpora}. In \bibinfo{booktitle}{\emph{ECIR'10: Advances in Information Retrieval}}, \bibfield{editor}{\bibinfo{person}{Cathal Gurrin}, \bibinfo{person}{Yulan He}, \bibinfo{person}{Gabriella Kazai}, \bibinfo{person}{Udo Kruschwitz}, \bibinfo{person}{Suzanne Little}, \bibinfo{person}{Thomas Roelleke}, \bibinfo{person}{Stefan R{\"u}ger}, {and} \bibinfo{person}{Keith {van Rijsbergen}}} (Eds.). \bibinfo{publisher}{{Springer Berlin Heidelberg}}, \bibinfo{address}{Berlin, Heidelberg}, \bibinfo{pages}{444--456}.
\newblock
\showISBNx{978-3-642-12275-0}


\bibitem[J\"{a}rvelin and Kek\"{a}l\"{a}inen(2002)]%
        {jarvelin.2002}
\bibfield{author}{\bibinfo{person}{Kalervo J\"{a}rvelin} {and} \bibinfo{person}{Jaana Kek\"{a}l\"{a}inen}.} \bibinfo{year}{2002}\natexlab{}.
\newblock \showarticletitle{Cumulated gain-based evaluation of IR techniques}.
\newblock \bibinfo{journal}{\emph{ACM Trans. Inf. Syst.}} \bibinfo{volume}{20}, \bibinfo{number}{4} (\bibinfo{date}{oct} \bibinfo{year}{2002}), \bibinfo{pages}{422–446}.
\newblock
\showISSN{1046-8188}
\href{https://doi.org/10.1145/582415.582418}{doi:\nolinkurl{10.1145/582415.582418}}


\bibitem[Li et~al\mbox{.}(2016)]%
        {Li.2016}
\bibfield{author}{\bibinfo{person}{Chenliang Li}, \bibinfo{person}{Haoran Wang}, \bibinfo{person}{Zhiqian Zhang}, \bibinfo{person}{Aixin Sun}, {and} \bibinfo{person}{Zongyang Ma}.} \bibinfo{year}{2016}\natexlab{}.
\newblock \showarticletitle{Topic Modeling for Short Texts with Auxiliary Word Embeddings}. In \bibinfo{booktitle}{\emph{SIGIR'16: Proceedings of the 39th International ACM SIGIR Conference on Research and Development in Information Retrieval}}, \bibfield{editor}{\bibinfo{person}{{Association for Computing Machinery}}} (Ed.). \bibinfo{publisher}{{Association for Computing Machinery}}, \bibinfo{address}{New York, NY, USA}, \bibinfo{pages}{165--174}.
\newblock
\showISBNx{9781450340694}
\href{https://doi.org/10.1145/2911451.2911499}{doi:\nolinkurl{10.1145/2911451.2911499}}


\bibitem[{Meilisearch}(2024)]%
        {meilisearchcharabia.2024}
\bibfield{author}{\bibinfo{person}{{Meilisearch}}.} \bibinfo{year}{2024}\natexlab{}.
\newblock \bibinfo{booktitle}{\emph{meilisearch/charabia}}.
\newblock
\urldef\tempurl%
\url{https://github.com/meilisearch/charabia}
\showURL{%
\tempurl}


\bibitem[Preiss(2012)]%
        {Preiss.2012}
\bibfield{author}{\bibinfo{person}{Judita Preiss}.} \bibinfo{year}{2012}\natexlab{}.
\newblock \showarticletitle{Identifying Comparable Corpora Using LDA}. In \bibinfo{booktitle}{\emph{NAACL HLT '12: Proceedings of the 2012 Conference of the North American Chapter of the Association for Computational Linguistics: Human Language Technologies}}, \bibfield{editor}{\bibinfo{person}{{Association for Computational Linguistics}}} (Ed.). \bibinfo{publisher}{{Association for Computational Linguistics}}, \bibinfo{address}{USA}, \bibinfo{pages}{558--562}.
\newblock
\showISBNx{9781937284206}
\urldef\tempurl%
\url{https://www.aclweb.org/anthology/N12-1065}
\showURL{%
\tempurl}


\bibitem[R\"{o}der et~al\mbox{.}(2015)]%
        {roedel.2015}
\bibfield{author}{\bibinfo{person}{Michael R\"{o}der}, \bibinfo{person}{Andreas Both}, {and} \bibinfo{person}{Alexander Hinneburg}.} \bibinfo{year}{2015}\natexlab{}.
\newblock \showarticletitle{Exploring the Space of Topic Coherence Measures}. In \bibinfo{booktitle}{\emph{Proceedings of the Eighth ACM International Conference on Web Search and Data Mining}} (Shanghai, China) \emph{(\bibinfo{series}{WSDM '15})}. \bibinfo{publisher}{Association for Computing Machinery}, \bibinfo{address}{New York, NY, USA}, \bibinfo{pages}{399–408}.
\newblock
\showISBNx{9781450333177}
\href{https://doi.org/10.1145/2684822.2685324}{doi:\nolinkurl{10.1145/2684822.2685324}}


\bibitem[S{\o}gaard et~al\mbox{.}(2015)]%
        {Sgaard.2015}
\bibfield{author}{\bibinfo{person}{Anders S{\o}gaard}, \bibinfo{person}{{\v{Z}}eljko Agi{\'c}}, \bibinfo{person}{H{\'e}ctor {Mart{\'i}nez Alonso}}, \bibinfo{person}{Barbara Plank}, \bibinfo{person}{Bernd Bohnet}, {and} \bibinfo{person}{Anders Johannsen}.} \bibinfo{year}{2015}\natexlab{}.
\newblock \showarticletitle{Inverted indexing for cross-lingual NLP}. In \bibinfo{booktitle}{\emph{ACL | IJCNLP'15: Proceedings of the 53rd Annual Meeting of the Association for Computational Linguistics and the 7th International Joint Conference on Natural Language Processing}}, \bibfield{editor}{\bibinfo{person}{Chengqing Zong} {and} \bibinfo{person}{Michael Strube}} (Eds.). \bibinfo{publisher}{{Association for Computational Linguistics}}, \bibinfo{address}{Stroudsburg, PA, USA}, \bibinfo{pages}{1713--1722}.
\newblock
\href{https://doi.org/10.3115/v1/P15-1165}{doi:\nolinkurl{10.3115/v1/P15-1165}}


\bibitem[Terragni et~al\mbox{.}(2020)]%
        {Terragni.2020}
\bibfield{author}{\bibinfo{person}{Silvia Terragni}, \bibinfo{person}{Elisabetta Fersini}, {and} \bibinfo{person}{Enza Messina}.} \bibinfo{year}{2020}\natexlab{}.
\newblock \showarticletitle{Constrained Relational Topic Models}.
\newblock \bibinfo{journal}{\emph{Information Sciences}}  \bibinfo{volume}{512} (\bibinfo{year}{2020}), \bibinfo{pages}{581--594}.
\newblock
\showISSN{00200255}
\href{https://doi.org/10.1016/j.ins.2019.09.039}{doi:\nolinkurl{10.1016/j.ins.2019.09.039}}


\bibitem[Vo(2022)]%
        {Vo.2022}
\bibfield{author}{\bibinfo{person}{Tham Vo}.} \bibinfo{year}{2022}\natexlab{}.
\newblock \showarticletitle{An Integrated Topic Modelling and Graph Neural Network for Improving Cross-Lingual Text Classification}.
\newblock \bibinfo{journal}{\emph{ACM Trans. Asian Low-Resour. Lang. Inf. Process.}} \bibinfo{volume}{22}, \bibinfo{number}{1}, Article \bibinfo{articleno}{22} (\bibinfo{date}{nov} \bibinfo{year}{2022}), \bibinfo{numpages}{18}~pages.
\newblock
\showISSN{2375-4699}
\href{https://doi.org/10.1145/3530260}{doi:\nolinkurl{10.1145/3530260}}


\bibitem[Wang et~al\mbox{.}(2020)]%
        {Wang.2020}
\bibfield{author}{\bibinfo{person}{Chaojie Wang}, \bibinfo{person}{Hao Zhang}, \bibinfo{person}{Bo Chen}, \bibinfo{person}{Dongsheng Wang}, \bibinfo{person}{Zhengjue Wang}, {and} \bibinfo{person}{Mingyuan Zhou}.} \bibinfo{year}{2020}\natexlab{}.
\newblock \showarticletitle{Deep Relational Topic Modeling via Graph Poisson Gamma Belief Network}. In \bibinfo{booktitle}{\emph{NeurIPS'20: Advances in Neural Information Processing Systems}}, \bibfield{editor}{\bibinfo{person}{{H. Larochelle}}, \bibinfo{person}{{M. Ranzato}}, \bibinfo{person}{{R. Hadsell}}, \bibinfo{person}{{M. F. Balcan}}, {and} \bibinfo{person}{{H. Lin}}} (Eds.), Vol.~\bibinfo{volume}{33}. \bibinfo{publisher}{{Curran Associates, Inc}}, \bibinfo{pages}{488--500}.
\newblock
\urldef\tempurl%
\url{https://proceedings.neurips.cc/paper/2020/file/05ee45de8d877c3949760a94fa691533-Paper.pdf}
\showURL{%
\tempurl}


\bibitem[Yang et~al\mbox{.}(2019)]%
        {Yang.2019}
\bibfield{author}{\bibinfo{person}{Weiwei Yang}, \bibinfo{person}{Jordan Boyd-Graber}, {and} \bibinfo{person}{Philip Resnik}.} \bibinfo{year}{2019}\natexlab{}.
\newblock \showarticletitle{A Multilingual Topic Model for Learning Weighted Topic Links Across Corpora with Low Comparability}. In \bibinfo{booktitle}{\emph{EMNLP | IJCNLP'19: Proceedings of the 2019 Conference on Empirical Methods in Natural Language Processing and the 9th International Joint Conference on Natural Language Processing}}, \bibfield{editor}{\bibinfo{person}{Kentaro Inui}, \bibinfo{person}{Jing Jiang}, \bibinfo{person}{Vincent Ng}, {and} \bibinfo{person}{Xiaojun Wan}} (Eds.). \bibinfo{publisher}{{Association for Computational Linguistics}}, \bibinfo{address}{Stroudsburg, PA, USA}, \bibinfo{pages}{1243--1248}.
\newblock
\href{https://doi.org/10.18653/v1/D19-1120}{doi:\nolinkurl{10.18653/v1/D19-1120}}


\bibitem[Zhao and Xing(2006)]%
        {Zhao.2006}
\bibfield{author}{\bibinfo{person}{Bing Zhao} {and} \bibinfo{person}{Eric~P. Xing}.} \bibinfo{year}{2006}\natexlab{}.
\newblock \showarticletitle{BiTAM: Bilingual Topic AdMixture Models for Word Alignment}. In \bibinfo{booktitle}{\emph{COLING | ACL'06: Proceedings of the COLING/ACL 2006 Main Conference Poster Sessions}}, \bibfield{editor}{\bibinfo{person}{{Association for Computational Linguistics}}} (Ed.). \bibinfo{publisher}{{Association for Computational Linguistics}}, \bibinfo{address}{Sydney, Australia}, \bibinfo{pages}{969--976}.
\newblock
\urldef\tempurl%
\url{https://www.aclweb.org/anthology/P06-2124}
\showURL{%
\tempurl}


\bibitem[Zhao et~al\mbox{.}(2017)]%
        {Zhao.2017}
\bibfield{author}{\bibinfo{person}{He Zhao}, \bibinfo{person}{Lan Du}, \bibinfo{person}{Wray Buntine}, {and} \bibinfo{person}{Gang Liu}.} \bibinfo{year}{2017}\natexlab{}.
\newblock \showarticletitle{{ MetaLDA: A Topic Model that Efficiently Incorporates Meta Information }}. In \bibinfo{booktitle}{\emph{2017 IEEE International Conference on Data Mining (ICDM)}}. \bibinfo{publisher}{IEEE Computer Society}, \bibinfo{address}{Los Alamitos, CA, USA}, \bibinfo{pages}{635--644}.
\newblock
\showISSN{2374-8486}
\href{https://doi.org/10.1109/ICDM.2017.73}{doi:\nolinkurl{10.1109/ICDM.2017.73}}


\end{thebibliography}

\end{document}